\documentclass[journal]{IEEEtran}
 
\usepackage[numbers]{natbib}
\usepackage{microtype}
\usepackage{graphicx}
\usepackage{booktabs} 
\usepackage{algorithm}
\usepackage{algorithmic}
\usepackage{hyperref}
\usepackage{multirow}
\usepackage{siunitx}
\usepackage{float}
\usepackage{bm}
\usepackage{hyperref}
\usepackage{tablefootnote}
\usepackage[table]{xcolor}  
\usepackage{colortbl}       

\usepackage{amsmath}
\usepackage{amssymb}
\usepackage{mathtools}
\usepackage{amsthm}
\usepackage{makecell}

\theoremstyle{plain}
\newtheorem{theorem}{Theorem}[section]

\newtheorem{corollary}[theorem]{Corollary}
\theoremstyle{definition}

\theoremstyle{remark}

\newcommand{\E}{\mathbb{E}}

\newcommand{\norm}[1]{\left\| #1 \right\|}

\newcommand{\mcal}[1]{\mathcal{#1}}

\newcommand{\wb}{\bm{w}}

\definecolor{myrowcolor}{RGB}{230, 242, 255}  
\definecolor{increasegreen}{RGB}{0, 150, 0}  

\begin{document}

\title{SpecGradFilter: A Spectral Gradient Filtering Framework for Taming Federated Heterogeneity}

\author{
    Liyang Yuan\thanks{
        Liyang Yuan is with the School of Artificial Intelligence,
        Jilin University, Changchun 130012, China.
        (e-mail: yuanly24@mails.jlu.edu.cn).
    },
    Yibo Yang\thanks{
        Yibo Yang is with the School of Artificial Intelligence,
        Shanghai Jiao Tong University, Shanghai, China,
        and also with King Abdullah University of Science and Technology (KAUST),
        Thuwal, Saudi Arabia.
        (e-mail: yibo.yang93@gmail.com).
    },
    Dandan Guo$^{*}$\thanks{
        Dandan Guo is with the School of Artificial Intelligence,
        Jilin University, Changchun 130012, China.
        (e-mail: guodandan@jlu.edu.cn).
        *Corresponding author.
    },
    Peter Richtárik\thanks{
        Peter Richtárik is with the Computer Science,
        King Abdullah University of Science and Technology (KAUST),
        Thuwal, Saudi Arabia.
        (e-mail: peter.richtarik@kaust.edu.sa).
    },
    Zhouchen Lin\thanks{
        Zhouchen Lin is with the State Key Lab of General Artificial Intelligence,
        School of Intelligence Science and Technology,
        Peking University, Beijing, China.
        (e-mail: zlin@pku.edu.cn).
    }
}

\maketitle

\begin{abstract}
  Federated Learning (FL) is fundamentally challenged by statistical heterogeneity, where non-identically distributed (non-IID) data induces ``client drift" that severely hampers global convergence. While existing approaches attempt to mitigate this drift through spatial-domain gradient correction or regularization, they overlook the intrinsic spectral structure of optimization signals. {In this work, we revisit client drift from a frequency-domain perspective and empirically observe a ``Spectral Bias of Drift": in the evaluated training trajectories, a disproportionate fraction of inter-client gradient disagreement is concentrated in the lower-frequency bands of the adopted structured parameter layout. Motivated by this observation, we propose SpecGradFilter, which suppresses a small low-frequency portion of selected layer gradients and can be implemented using an FFT-based projection or spatial-domain detrending approximations. For the FFT-based realization, we establish a conditional convergence guarantee when the projection retains sufficient global descent signal and the post-filtering client heterogeneity is bounded. Extensive experiments on CIFAR-10/100, Tiny-ImageNet, and additional benchmarks show improved accuracy and time-to-target in the tested non-IID settings without increasing the per-round communication payload.}
\end{abstract}


\section{Introduction}
\IEEEPARstart{F}{ederated} Learning (FL) \cite{fedavg} has emerged as a promising paradigm for enabling collaborative machine learning across decentralized clients without sharing raw data, thereby preserving data privacy \cite{efficient1,efficient2,efficient3,yang2026co}. Despite its appeal, the performance of FL is fundamentally hindered by statistical heterogeneity across clients. When local datasets are non-IID, the local optimization objectives deviate from the global objective, causing client updates toward client-specific optima. This phenomenon, commonly referred to as client drift, significantly slows down convergence and degrades the generalization performance of the global model \cite{Khaled2019TighterTF,fl_challenge1,fl_challenge2,fl_challenge3,feddyn,moon,shi2025towards}.

To mitigate client drift, existing research has predominantly focused on spatial-domain optimization strategies. Representative approaches include constraining local update magnitudes through proximal regularization (e.g., FedProx) \cite{fedprox}, correcting update directions using control variates (e.g., SCAFFOLD) \cite{scaffold}, dynamically aligning global and local objectives via regularization mechanisms (e.g., FedDyn) \cite{feddyn}, as well as optimizing federated aggregation by dynamically weighing local updates based on dataset size and category distribution discrepancy (e.g., FedDisco) \cite{feddisco} or by leveraging client-level representations to adjust aggregation weights (e.g., FedAWA) \cite{fedawa}. 
While these methods have demonstrated effectiveness under certain conditions, they fundamentally treat gradients as unstructured numerical vectors in the parameter space. As a result, they implicitly assume that all gradient components contribute equally, overlooking potential structural patterns and spectral correlations within the gradients themselves.

In this work, we revisit the client drift problem from a novel frequency-domain perspective. Our investigation is inspired by the well-established spectral bias of neural networks \cite{rahaman2018spectral,xu2019frequency}, which suggests that deep models tend to learn low-frequency (smooth) functions before capturing high-frequency details. {Here, frequency refers to the rFFT spectrum of gradients flattened under a specified parameter layout, rather than directly to the input--output frequency of the learned function. We empirically observe that, in the evaluated ResNet-20/CIFAR-10 trajectories, inter-client gradient disagreement is nonuniformly distributed across this spectrum, with lower-frequency bands carrying a disproportionate fraction of the measured heterogeneity energy. We term this empirical pattern the Spectral Bias of Drift; it characterizes the spectral allocation of client disagreement without assigning an exclusive semantic meaning to either subspace.}

This observation reveals a fundamental limitation of standard federated optimization schemes such as FedAvg, which aggregate client model updates without distinguishing their underlying spectral components. Since local updates implicitly encode gradient information across all frequencies, aggregating updates dominated by highly inconsistent low-frequency components inadvertently amplifies client drift and hinders global convergence. Motivated by this insight, we propose SpecGradFilter, a spectral framework for federated optimization that mitigates heterogeneity by  suppressing divergent low-frequency components during local updates. 
Unlike gradient compression techniques designed primarily for communication efficiency \cite{fedfd}, 
{SpecGradFilter suppresses lower-frequency components that contain disproportionate client disagreement in our measurements while retaining the complementary subspace. Its FFT-based projection can also be approximated by spatial detrending operators, and the filter is applied to architecture-appropriate layers whose parameter layouts preserve meaningful structural locality.}
This flexibility enables SpecGradFilter to be easily integrated into existing federated learning pipelines {without changing the communication payload, while introducing additional local computation}. Our main contributions are summarized as follows:
\begin{itemize}
    \item  {We uncover the \emph{Spectral Bias of Drift} in the evaluated FL trajectories and provide empirical evidence that lower-frequency bands of the adopted structured parameter layout carry a disproportionate fraction of client-gradient disagreement.}
    
    \item  We propose {SpecGradFilter}, a simple yet effective framework that mitigates client drift by suppressing divergent low-frequency gradient components, with implementations in both frequency and spatial domains.
    
    \item Extensive experiments  under highly Non-IID settings demonstrate that SpecGradFilter consistently 
     achieves superior performance and convergence speed compared to a wide range of strong federated baselines.    
    
\end{itemize}

\section{Related Work}
\subsection{Federated Learning under Data Heterogeneity}
Federated Learning (FL) faces fundamental optimization challenges under data heterogeneity. Non-IID distributions induce client drift, where local updates diverge from the global trajectory due to biased objectives and multi-step local training. {This effect, exacerbated by limited client participation, can substantially hamper convergence and generalization} \cite{fedavg,fl_challenge1,fl_challenge2,fedpe,fedscpack,Gao_2026_CVPR}.

Existing approaches to mitigating client drift can be broadly categorized into local objective regularization and modified aggregation strategies. Local regularization aims to align local updates with the global objective. Representative works include FedProx \cite{fedprox}, which introduces an $\ell_2$ proximal term to constrain model deviation, and FedDyn \cite{feddyn}, which employs dynamic regularization for update consistency. Alternatively, SCAFFOLD \cite{scaffold} utilizes control variates to explicitly correct gradient bias. Recent refinements further explore auxiliary variables to track parameter discrepancies (e.g., FedDC \cite{feddc}) or calibrate loss functions to address label skew (e.g., FedLC \cite{fedlc}). Another line of research optimizes server-side aggregation. Methods such as FedLAW \cite{fedlaw}, FedAWA \cite{fedawa}, and FedDisco \cite{feddisco}  redesign aggregation weights by accounting for data imbalance, client representations, or distribution discrepancies. Furthermore, InCo  \cite{inco} leverages cross-layer information to enhance deeper layer alignment without extra communication. While these approaches are effective, most of them operate in the parameter or gradient value space, emphasizing magnitude, direction, or statistical weighting of updates. In contrast, relatively little attention has been paid to the structural properties of gradients themselves, particularly from a frequency-domain perspective, which motivates our work.

\subsection{From Spectral Bias to Frequency-Domain Optimization}
The spectral behavior of deep neural networks has been extensively studied under the framework of spectral bias or the Frequency Principle. Early theoretical and empirical analyses show that neural networks tend to learn target functions from low-frequency components to high-frequency ones during training \cite{rahaman2018spectral,xu2019frequency}. Lower-frequency components are not only learned earlier but are also more robust to parameter perturbations, revealing an implicit bias in deep optimization dynamics. However, we argue that in federated contexts, this rapid capture of low-frequency information can be double-edged, as these components are also most susceptible to absorbing discordant, client-specific distributional biases.

Several recent works have introduced frequency-domain techniques into federated or distributed learning, primarily motivated by communication efficiency. For instance, SuperNeurons \cite{wang2018super} and FedFD  \cite{fedfd} apply FFT-based filtering or sparsification to reduce transmission costs by targeting low-energy or noisy components. Similarly, FedFT \cite{fedft}  utilizes the Discrete Cosine Transform (DCT) for compact parameter representation and flexible updates. Despite their effectiveness, these methods are largely designed as compression or acceleration techniques, where frequency filtering is guided by energy or sparsity considerations rather than optimization stability. In contrast, our method leverages frequency-domain filtering as a regularization mechanism to directly mitigate client drift. By suppressing energy-dominant yet drift-prone low-frequency signals, we prioritize optimization alignment over mere data compression.

\begin{figure}[t]
    \centering
    \includegraphics[width=0.45\textwidth]{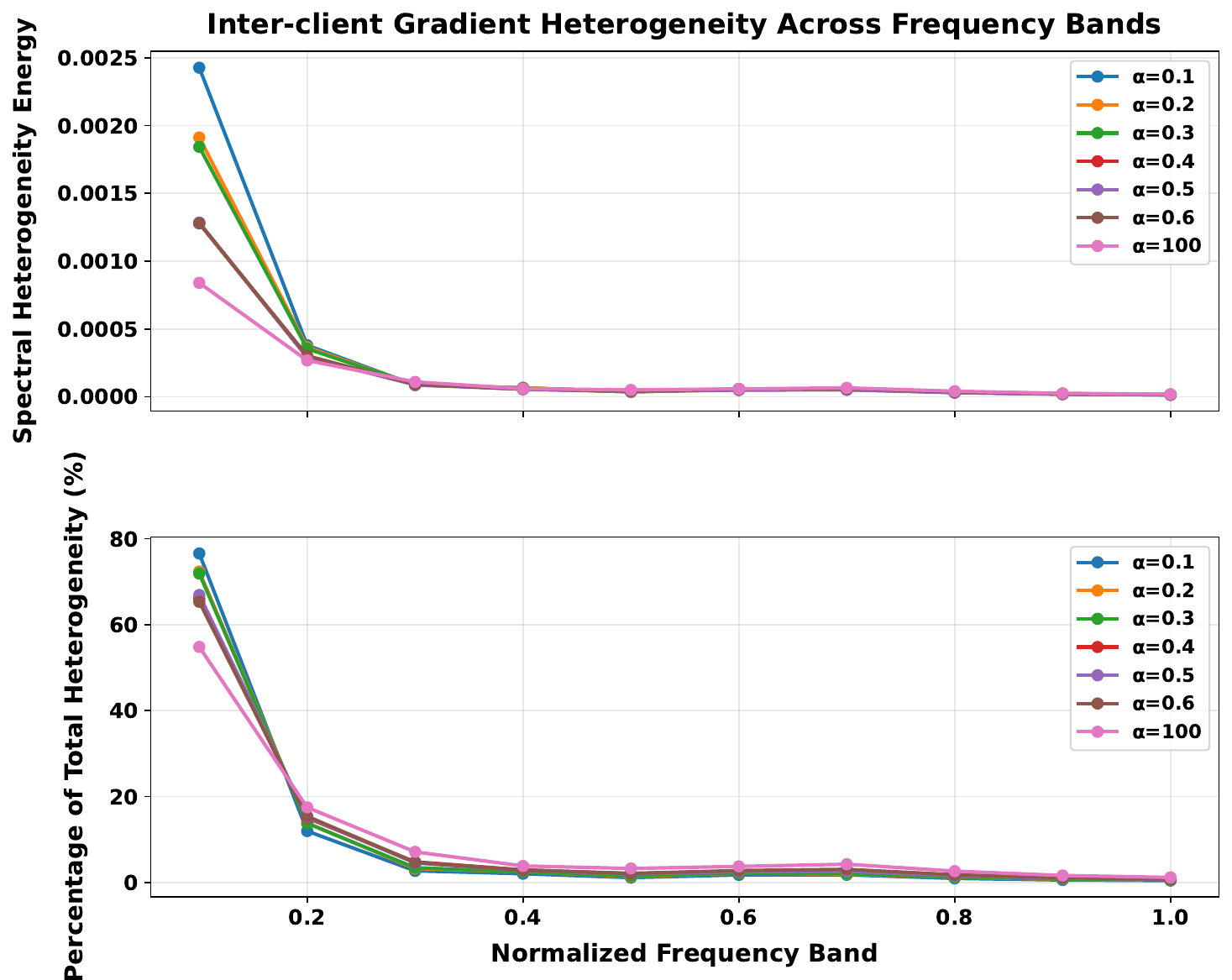}

    \caption{{\small{
        Spectral distribution of {active-client} gradient heterogeneity under different Dirichlet data partitions. Top: mean spectral heterogeneity energy in each frequency band, measuring the absolute magnitude of client gradient disagreement relative to the {within-round active-client consensus}. Bottom: percentage of the total heterogeneity energy contributed by each frequency band. Results are obtained using ResNet-20 on CIFAR-10 {with participation ratio $0.1$}.
    }}}

    \label{fig:grad_divergence}
\end{figure}

\section{Understanding Client Drift via Spectral Lens in Federated Learning}

\subsection{Preliminaries}
Federated Learning is a distributed learning paradigm that enables collaborative model training over data from $K$ clients without centralizing their private datasets $\mathcal{D}_{1:K}$. A central server coordinates the optimization process by aggregating model updates from participating clients, whose learning objective is to minimize the global empirical risk.
{Let $n_k = |\mathcal{D}_k|$ denote the number of training samples on client $k$, and let $N = \sum_{k=1}^{K} n_k$ denote the total number of training samples across all clients. {Define $p_k:=n_k/N$, so that $\sum_{k=1}^{K}p_k=1$.}}
\begin{equation}\label{eq:fl_objective}
\begin{aligned}
{
F(w)} 
&= {
\sum_{k=1}^{K} \frac{n_k}{N} {F_k(w)}}, \\
{F_k(w)}
&= {\frac{1}{n_k}}
\sum_{(x,y) \in \mathcal{D}_k} 
\mathcal{L}_k(w; x, y),
\end{aligned}
\end{equation}
where {$F_k(w)$} denotes the local objective defined on client $k$'s data, $w$ means the to-be-learned parameter. The most widely adopted optimization algorithm for FL is FedAvg~\citep{fedavg}, which proceeds in iterative communication rounds. In $t$-th round, the server {(1) broadcasts} the current global model $w^t$ to a subset of clients {$S_t$}. The $k$-th client then performs several steps of {(2) local training} on its private data, producing an updated local model $w_k^{t+1}$. These local models are subsequently {(3) uploaded} to the server and {(4) aggregated} to form the next global model:
{
\[
w^{t+1}
=
\sum_{k \in S_t}
\frac{n_k}{\sum_{j \in S_t} n_j}
w_k^{t+1}.
\]
}

Despite its success, the fundamental challenge in FL is {client drift}. Mathematically, let {$g_k = \nabla F_k(w)$} be the local gradient and $g = \nabla F(w)$ be the global gradient. Due to the Non-IID nature of local datasets $\mathcal{D}_k$, the local update direction often deviates from the global optimum:
\begin{equation}
\Delta_k = \| g_k - g\|,
\end{equation}
where $\Delta_k$ represents the magnitude of the drift. Prior research typically views $\Delta_k$ as an unstructured numerical error in the spatial domain. However, this perspective overlooks the internal structural properties of the model parameters. We raise a  question: \textit{Is this drift uniformly distributed across the internal components of the parameter vector? If we transform this deviation into the frequency domain, what patterns will emerge?}

{
\subsection{Our Observation: Spectral Bias of Drift}
\label{section:observation}

To investigate the frequency-domain structure of client drift, we interpret the flattened gradient vector
$\mathbf{g}_{k,l}^{(t)} \in \mathbb{R}^{d_l}$ of client $k$ at layer $l$ and communication round $t$ as a discrete signal. We then project it into the frequency domain using the real-valued Discrete Fourier Transform (rFFT):
\begin{equation}
\widehat{\mathbf{g}}_{k,l}^{(t)}
=
\mathcal{F}\!\left(\mathbf{g}_{k,l}^{(t)}\right)
\in
\mathbb{C}^{\lfloor d_l/2 \rfloor+1},
\label{eq:rfft}
\end{equation}
where the orthonormal normalization is adopted so that the spatial- and frequency-domain energies satisfy Parseval's identity.
{The diagnostic follows the actual partial-participation
training protocol.  Let $S_t$ denote the clients active at checkpoint $t$,
with $|S_t|/K=0.1$, and define their round-normalized aggregation weights by
\begin{equation}
    p_{k,t}:=
    \frac{n_k}{\sum_{j\in S_t}n_j},
    \qquad k\in S_t,
    \qquad \sum_{k\in S_t}p_{k,t}=1.
    \label{eq:diagnostic_active_weights}
\end{equation}
}
For each layer, we first construct the spectral representation of the
{active-client consensus gradient} as
\begin{equation}
\overline{\widehat{\mathbf{g}}}_{l}^{(t)}
=
{\sum_{k\in S_t} p_{k,t}}
\widehat{\mathbf{g}}_{k,l}^{(t)},
\label{eq:mean_spectrum}
\end{equation}
{where $p_{k,t}$ is the same within-round sample-size weight
used by the aggregation rule.} Since the Fourier transform is linear,
$\overline{\widehat{\mathbf{g}}}_{l}^{(t)}$ is equivalent to the rFFT of the
{active-client weighted-average gradient}.

We uniformly partition the one-sided spectrum into {$B_{\mathrm{freq}}$} disjoint frequency bands
{$\{\mathcal{B}_m\}_{m=1}^{B_{\mathrm{freq}}}$}. For each band, we define the spectral heterogeneity energy as
\begin{equation}
H_{l,m}^{(t)}
=
{\sum_{k\in S_t} p_{k,t}}
\sum_{q\in\mathcal{B}_m}
\omega_{l,q}
\left|
\widehat{\mathbf{g}}_{k,l}^{(t)}[q]
-
\overline{\widehat{\mathbf{g}}}_{l}^{(t)}[q]
\right|^2,
\label{eq:band_heterogeneity_energy}
\end{equation}
where $\omega_{l,q}$ is the Parseval correction weight for the one-sided rFFT spectrum. Specifically, the DC component and, when present, the Nyquist component are assigned weight $1$, whereas the remaining positive-frequency components are assigned weight $2$. This correction accounts for the omitted negative-frequency coefficients of a real-valued signal.

Unlike a pairwise distance metric, $H_{l,m}^{(t)}$ directly measures the weighted variance of client gradients around their {active-client consensus} within frequency band $m$. More importantly, the resulting band energies are additive:
\begin{equation}
{\sum_{m=1}^{B_{\mathrm{freq}}}} H_{l,m}^{(t)}
=
{\sum_{k\in S_t}p_{k,t}}
\left\|
\mathbf{g}_{k,l}^{(t)}
-
\overline{\mathbf{g}}_{l}^{(t)}
\right\|_2^2,
\label{eq:parseval_heterogeneity}
\end{equation}
where
$\overline{\mathbf{g}}_{l}^{(t)}
=
{\sum_{k\in S_t}p_{k,t}\mathbf{g}_{k,l}^{(t)}}$.
Therefore, the sum of the frequency-band heterogeneity energies exactly
recovers the total {active-client} gradient heterogeneity of
the corresponding layer.

We then average the band-wise energy across the {diagnostic layer set $\mathcal{S}_{\mathrm{diag}}$}:
\begin{equation}
H_m^{(t)}
=
{\frac{1}{|\mathcal{S}_{\mathrm{diag}}|}
\sum_{l\in\mathcal{S}_{\mathrm{diag}}}}
H_{l,m}^{(t)}.
\label{eq:round_band_energy}
\end{equation}
Each valid layer is assigned equal weight in this aggregation. We retain weight tensors shared by all participating clients, while excluding bias parameters, normalization parameters, downsampling layers, and tensors whose lengths are insufficient to support the prescribed frequency partition.

To characterize the spectral pattern over training, we collect gradients from a set of communication rounds {$\mathcal{T}$} and compute the mean heterogeneity energy of each band:
\begin{equation}
\overline{H}(m)
=
{\frac{1}{|\mathcal{T}|}
\sum_{t\in\mathcal{T}}}
H_m^{(t)}.
\label{eq:mean_band_energy}
\end{equation}
We further calculate the fraction of the total heterogeneity energy contributed by each frequency band:
\begin{equation}
{\varphi_m}
=
\frac{\overline{H}(m)}
{{\sum_{b=1}^{B_{\mathrm{freq}}}\overline{H}(b)}}.
\label{eq:band_energy_share}
\end{equation}
By construction, {$\sum_{m=1}^{B_{\mathrm{freq}}}\varphi_m=1$}. The absolute energy
$\overline{H}(m)$ characterizes the magnitude of client heterogeneity in each frequency band, whereas {$\varphi_m$} characterizes how the total heterogeneity is distributed over the spectrum. 

We conduct experiments on CIFAR-10 using ResNet-20 under Dirichlet data partitions with
$\alpha\in\{0.1,0.2,0.3,0.4,0.5,0.6,100\}$, where smaller $\alpha$ indicates stronger statistical heterogeneity and $\alpha=100$ approximates a near-IID setting. Client gradients are collected at communication rounds
{$\mathcal{T}=\{0,30,60,90\}$} {from the
clients active in the corresponding rounds of training with participation
ratio $0.1$}. The gradients are flattened layer-wise, transformed using an
orthonormal rFFT, and uniformly divided into
{$B_{\mathrm{freq}}=10$} frequency bands. The band-wise heterogeneity energy is averaged equally across valid layers and the selected rounds. We also numerically verify the Parseval-consistent decomposition, obtaining a maximum error of only
$3.73\times10^{-9}$.

As shown in Figure~\ref{fig:grad_divergence}, the upper panel reports the absolute heterogeneity energy $\overline{H}(m)$, while the lower panel reports its percentage contribution {$100\varphi_m$}. {In this ResNet-20/CIFAR-10 active-client diagnostic, gradient heterogeneity consistently exhibits a disproportionate concentration in the lower-frequency region of the adopted layout.} In the near-IID setting with $\alpha=100$, the first frequency band accounts for $50.83\%$ of the total heterogeneity energy, whereas this proportion increases to $71.61\%$ under $\alpha=0.1$. The first two bands jointly account for $67.45\%$ and $82.37\%$ under these two settings, respectively.

Increasing statistical heterogeneity also amplifies the lowest-frequency components more strongly. As $\alpha$ decreases from $100$ to $0.1$, the total heterogeneity energy increases by approximately $2.55\times$, while the energy of the first band increases by $3.59\times$ and contributes about $79.7\%$ of the total increase. In comparison, the aggregated energy of the last five bands increases by only $1.44\times$. These results reveal a clear \textbf{Spectral Bias of Drift}: {in the evaluated active-client trajectories, gradient disagreement is disproportionately concentrated in the lower-frequency region, and the additional disagreement associated with stronger non-IID partitioning accumulates primarily in this region. This observation motivates suppressing a small low-frequency band while retaining the complementary subspace.} To further examine the immediate effect of spectral
suppression, we apply the high-pass operator to the client gradients and measure the remaining inter-client divergence. As reported in the supplementary material Figure 2, low-frequency suppression consistently removes a substantial portion of client disagreement. This diagnostic motivates the proposed local filtering procedure.
}

\section{ METHODOLOGY}

\begin{figure*}[t] 
    \centering
    \includegraphics[width=0.8\textwidth]{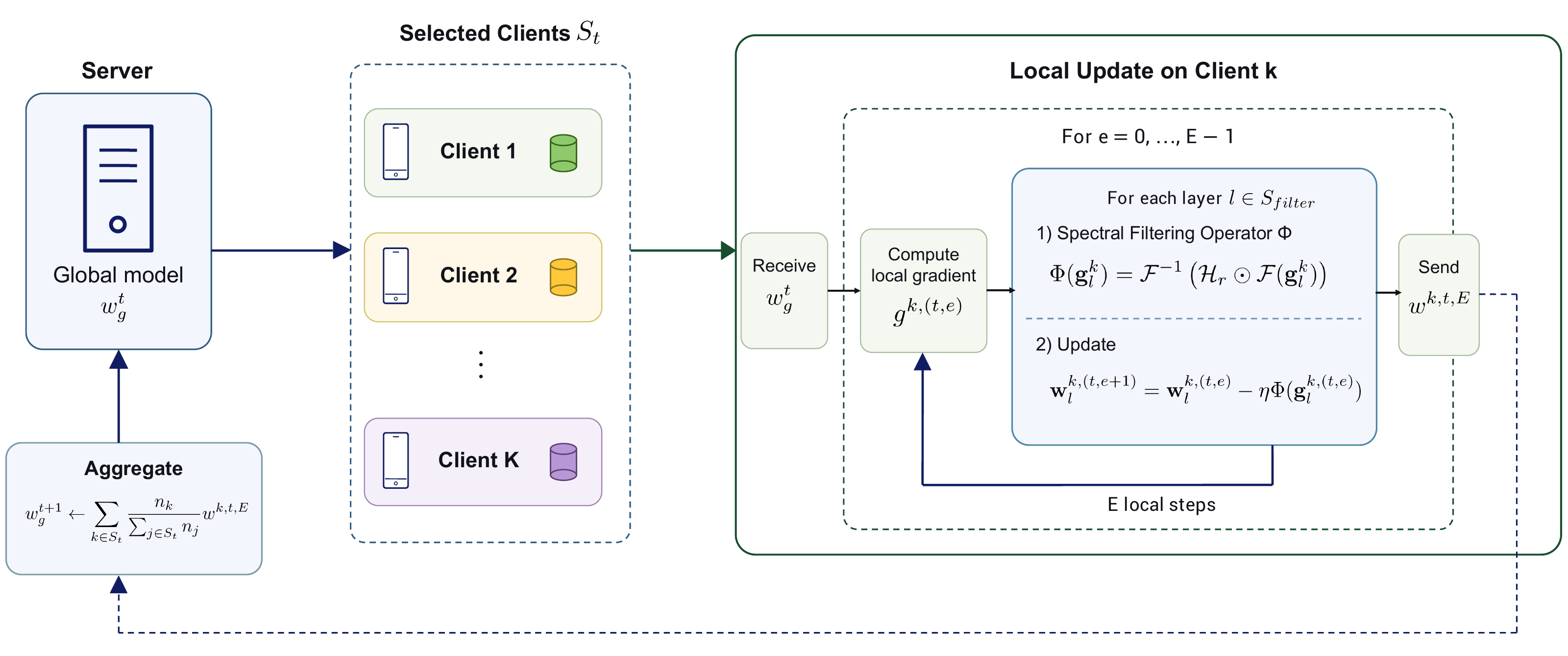}
    \caption{
    {
    Overview of the proposed SpecGradFilter framework integrated with FedAvg. 
    The figure illustrates the overall training pipeline including local spectral 
    gradient filtering, client updates, and server-side aggregation.
    }
    }
    \label{fig:model}
\end{figure*}

Motivated by the Spectral Bias of Drift observed in above Section, we propose SpecGradFilter, a unified framework designed to tame federated heterogeneity by selectively suppressing drift-prone spectral components. The core of our framework is the spectral modulation operator $\Phi$, which serves as a structured regularizer to align local updates across clients. The overall workflow of SpecGradFilter integrated with FedAvg is illustrated in Fig.~\ref{fig:model}.

\subsection{Realizations of the Spectral Filtering Operator}
\label{section:filter-method}

While $\Phi$ represents a generic high-pass operation, its most precise realization is achieved in the frequency domain. Let $\mathbf{g}_l^k \in \mathbb{R}^{d_l}$ denote the flattened gradient of client $k$ at layer $l$. The spectral filtering process is formally defined as:

\begin{equation}
\Phi(\mathbf{g}_l^k) = \mathcal{F}^{-1} \left( \mathcal{H}_r \odot \mathcal{F}(\mathbf{g}_l^k) \right),\label{eq:spec_filter}
\end{equation}

where $\mathcal{F}$ and $\mathcal{F}^{-1}$ denote the real-valued Discrete Fourier Transform (rFFT) and its inverse, respectively. $\mathcal{H}_r$ is a spectral mask characterized by the suppression ratio $r \in [0, 1)$. {The cutoff is computed separately from each layer's rFFT length, so the same ratio suppresses a proportional number of coefficients in every selected layer.}

Following our empirical findings that drift predominantly resides in low frequencies, we instantiate $\mathcal{H}_r$ as a high-pass filter. Specifically, let $\mathbf{z} = \mathcal{F}(\mathbf{g}_l^k) \in \mathbb{C}^{\lfloor d_l/2 \rfloor + 1}$ be the vector of spectral coefficients. The mask $\mathcal{H}_r \in \{0, 1\}^{\lfloor d_l/2 \rfloor + 1}$ is defined to zero out the first $\lfloor r \cdot (\lfloor d_l/2 \rfloor + 1) \rfloor$ low-frequency components:
\begin{equation}
[\mathcal{H}_r]_m =
\begin{cases}
0, & m < \lfloor r \cdot (\lfloor d_l/2 \rfloor + 1) \rfloor, \\[2mm]
1, & \text{otherwise},
\end{cases}
\label{eq:mask}
\end{equation}
where {$m\in\{0,\ldots,\lfloor d_l/2\rfloor\}$} denotes the frequency index. {Applying this spectral mask removes the selected low-frequency coefficients and retains the complementary spectral subspace.} {If the cutoff evaluates to zero, the mask is the identity and the layer is effectively unfiltered.}

\textbf{Local Update with Spectral Filtering.} During each communication round $t$, once a client $k$ is sampled, it performs {$E$ local stochastic-gradient steps}. To rectify the optimization trajectory in real-time, the spectral operator $\Phi$ is applied to the local gradient at each step. Formally, for each layer $l$, the parameters are updated as follows:
\begin{equation}\mathbf{w}_l^{k, (t, e+1)} = \mathbf{w}_l^{k, (t, e)} - \eta \Phi(\mathbf{g}_l^{k, (t, e)}), 
\label{eq:update}
\end{equation}

where $\eta$ is  local learning rate and $\mathbf{w}_l^{k,(t,0)} \!=\! \mathbf{w}_l^{(t)}$ is the global model parameter received from the server. 
{Intuitively, applying $\Phi$ in the local update loop suppresses the selected low-frequency band that carries substantial client disagreement in our measurements. The module can be combined with several FL optimizers using the architecture-specific target layers and optimizer-specific integration rules described in the supplementary material.} Please see our algorithm in Algorithm~\ref{alg:specgradfilter}.

\textbf{Generalizing to Spatial-Domain Realizations.} To demonstrate the versatility of our spectral filtering principle, we provide alternative realizations of $\Phi$ that operate directly in the spatial domain. Inspired by signal processing, the low-frequency components (representing the "trend" or client-specific bias) can be effectively captured via local smoothing kernels without explicit frequency transformation. Specifically, we instantiate $\Phi$ using Local Average Pooling Detrend (LAP-D) or Gaussian Detrend (GD), defined as $\Phi(\mathbf{g}_l^k) = \mathbf{g}_l^k - \text{Smooth}(\mathbf{g}_l^k)$. Here, $\text{Smooth}(\cdot)$ acts as a spatial low-pass filter that extracts the coarse-grained drift, where details are provided in Section I of the supplementary material. {By subtracting this trend, these variants produce an approximate high-pass effect analogous to, but not mathematically equivalent to, the hard frequency-domain mask.} This provides a unified spatial-domain optimization pipeline, which is particularly advantageous for hardware environments where convolution primitives are more optimized than Fourier transforms, while ensuring strict linear-time complexity $\mathcal{O}(d_l)$.

\begin{algorithm}[t]
\caption{SpecGradFilter with FedAvg}
\label{alg:specgradfilter}
\begin{algorithmic}[1]
\STATE {\bfseries Input:} Communication rounds $T$, {local stochastic-gradient steps $E$},
local learning rate $\eta$, low-frequency filtering ratio $r$,
{target layer set $\mathcal{S}_{\mathrm{filter}}$}
\STATE {\bfseries Output:} Global model $w_g^T$
\STATE Initialize global model $w_g^0$
\STATE {
Let $n_k = |\mathcal{D}_k|$ denote the number of local training
samples on client $k$
}

\FOR{$t = 0$ {\bfseries to} $T-1$}
    \STATE Randomly select active client set $S_t$

    \FOR{each client $k \in S_t$ {\bfseries in parallel}}
        \STATE $w^{k,t,0} \gets w_g^t$

        \FOR{$e = 0$ {\bfseries to} $E-1$}
            \STATE Compute local gradient
            $g^{k,t,e} \gets \nabla \mathcal{L}_k(w^{k,t,e})$

            \FOR{each layer $l$ in the model}
                \IF{{$l \in \mathcal{S}_{\mathrm{filter}}$}}
                    \STATE {
                    Compute the filtered gradient
                    $\Phi(\mathbf{g}_l^{k,t,e})$
                    via Eq.~\eqref{eq:spec_filter}
                    }
                    \STATE {
                    Update $\mathbf{w}_l^{k,t,e+1}$
                    via Eq.~\eqref{eq:update}
                    }
                \ELSE
                    \STATE {
                    $\mathbf{w}_l^{k,t,e+1}
                    \gets
                    \mathbf{w}_l^{k,t,e}
                    - \eta \mathbf{g}_l^{k,t,e}$
                    }
                \ENDIF
            \ENDFOR
        \ENDFOR

        \STATE Send $w^{k,t,E}$ to the server
    \ENDFOR

    \STATE {
    $
    w_g^{t+1}
    \gets
    \displaystyle
    \sum_{k \in S_t}
    \frac{n_k}
    {\sum_{j \in S_t} n_j}
    w^{k,t,E}
    $
    }

\ENDFOR

\STATE {\bfseries return} $w_g^T$
\end{algorithmic}
\end{algorithm}

\subsection{The Spectral Preconditioning View}

The proposed frequency-domain filtering can also be interpreted from a preconditioning perspective. 
Specifically, the rFFT-based suppression constitutes a linear projection in the spectral Hilbert space. Let  $\mathbf{D} = \text{diag}(\mathcal{H}_r)$ be the diagonal mask matrix constructed from the spectral mask defined in Eq.~\ref{eq:mask}. The filtered gradient can be equivalently written as:
\begin{equation}
\tilde{\mathbf{g}} = \mathcal{F}^{-1} \mathbf{D} \mathcal{F} (\mathbf{g}) = {\mathbf{P}_{\mathrm{spec}}} \mathbf{g},
\end{equation}
where {$\mathbf{P}_{\mathrm{spec}}$} $= \mathcal{F}^{-1} \mathbf{D} \mathcal{F}$ is the spectral projection preconditioner. Unlike classical preconditioners that aim to correct curvature-induced ill-conditioning (e.g., second-order or adaptive methods), {$\mathbf{P}_{\mathrm{spec}}$} targets ill-conditioning arising from statistical heterogeneity across clients. {As shown in our empirical analysis, heterogeneity-induced gradient discrepancies are highly anisotropic and contain a substantial, disproportionate amount of energy in the suppressed low-frequency band, with layer-dependent strength. Projecting gradients away from this band therefore reduces inter-client disagreement without requiring curvature estimation.}

{Because every client uses the same layer-wise mask, the
full-model FFT operator $P^{(\mathrm{high})}$ is a client-independent
linear orthogonal projector: $(P^{(\mathrm{high})})^\top=P^{(\mathrm{high})}$
and $(P^{(\mathrm{high})})^2=P^{(\mathrm{high})}$.  Consequently it
commutes with weighted aggregation,
$P^{(\mathrm{high})}\sum_{k=1}^{K}p_k g_k
=\sum_{k=1}^{K}p_kP^{(\mathrm{high})}g_k$.  On layers that are not
filtered, its corresponding block is the identity.}

{
\subsection{Theoretical Analysis}

{We provide a conditional convergence guarantee for the FFT-based orthogonal-projection realization of SpecGradFilter and show that its heterogeneity-dependent error is controlled by the client disagreement remaining after spectral filtering. The theorem requires the retained subspace to preserve a sufficient fraction of the global descent signal and is parameterized by the heterogeneity remaining after spectral filtering. The complete assumptions, structural sufficient conditions, and proofs are provided in Section~II of the supplementary material.}

\begin{theorem}[Convergence of SpecGradFilter]
\label{thm:main}
Suppose that the local objectives are $L$-smooth, the stochastic gradients are unbiased with bounded variance, {the gradient noises are conditionally independent across clients and form martingale-difference sequences across local steps, $F$ is bounded below,} and the retained global gradient satisfies
\begin{equation}
    \norm{
        P^{(\mathrm{high})}\nabla F(\wb)
    }^2
    \geq
    (1-\rho)
    \norm{
        \nabla F(\wb)
    }^2,
    \qquad
    \rho\in[0,1).
\end{equation}
Assume weighted full client participation and the retained
heterogeneity bound
\begin{equation}
\begin{aligned}
    &\sum_{k=1}^{{K}}p_k
    \norm{
        P^{(\mathrm{high})}
        \left(
            \nabla F_k(\wb)-\nabla F(\wb)
        \right)
    }^2
    \leq
    \delta_{\mathrm{high}}^2.
\end{aligned}
\label{eq:main_retained_heterogeneity}
\end{equation}
{
Here, $\delta_{\mathrm{high}}^2$ is any uniform upper bound on
the heterogeneity retained by the actual update operator.  The
convergence proof below uses this retained budget directly; its
general concentration-based and stronger layout-smooth
specializations are stated after the theorem.
}

Let $E$ be the number of local update steps per round. For
$T\geq64$ and
\begin{equation}
    \eta
    =
    \frac{1}{LE\sqrt{T}},
\end{equation}
SpecGradFilter satisfies
\begin{equation}
\begin{aligned}
    \frac{1}{T}
    \sum_{t=0}^{T-1}
    \E\!\left[
        \norm{
            \nabla F(\wb^t)
        }^2
    \right]
    &\leq
    \frac{
        8L\left(
            F(\wb^0)-F^*
        \right)
    }{
        (1-\rho)\sqrt{T}
    }\\
    &\quad+
    \frac{
        36\delta_{\mathrm{high}}^2
        +
        12\sigma_{\mathrm{loc}}^2
    }{
        (1-\rho)T
    }\\
    &\quad+
    \frac{
        8\bar{\sigma}^2
    }{
        E(1-\rho)\sqrt{T}
    },
\end{aligned}
\label{eq:main_convergence}
\end{equation}
where
\begin{equation}
    \sigma_{\mathrm{loc}}^2
    :=
    \sum_{k=1}^{{K}}p_k\sigma_k^2,
    \qquad
    \bar{\sigma}^2
    :=
    \sum_{k=1}^{{K}}p_k^2\sigma_k^2,
\end{equation}
and $F^*:=\inf_{\wb}F(\wb)$.
\end{theorem}

{
\paragraph{From spectral concentration to retained heterogeneity.}
For a filtered layer with nonzero total client disagreement, define
\begin{equation}
    \kappa_l(\wb)
    :=
    \frac{H_l^{(\mathrm{low})}(\wb)}
    {H_l^{(\mathrm{low})}(\wb)
     +H_l^{(\mathrm{high})}(\wb)}.
    \label{eq:main_kappa_definition}
\end{equation}
Because the low- and high-frequency subspaces are orthogonal,
\begin{equation}
    H_l^{(\mathrm{high})}(\wb)
    =
    \bigl(1-\kappa_l(\wb)\bigr)H_l(\wb),
    \qquad
    H_l=H_l^{(\mathrm{low})}+H_l^{(\mathrm{high})}.
    \label{eq:main_exact_retained_identity}
\end{equation}
Suppose
$\kappa_l(\wb)\geq\underline{\kappa}\in[0,1]$ over the filtered
layer set $\mcal{S}_{\mathrm{filter}}$, its total heterogeneity
satisfies
$H_{\mcal{S}}(\wb)\leq\delta_{\mcal{S}}^2$, and the heterogeneity of
unfiltered layers is at most $\delta_{\mathrm{res}}^2$.  Then an
admissible budget in Eq.~\eqref{eq:main_retained_heterogeneity} is
\begin{equation}
    \boxed{
    \delta_{\mathrm{high}}^2
    =
    (1-\underline{\kappa})\delta_{\mcal{S}}^2
    +\delta_{\mathrm{res}}^2}.
    \label{eq:main_derived_high_heterogeneity}
\end{equation}
Thus any $\underline{\kappa}>0$ certifies a strict reduction of the
filtered-set heterogeneity; it is unnecessary to require
$H_l^{(\mathrm{low})}>H_l^{(\mathrm{high})}$.

The supplementary analysis also provides a structural sufficient
condition. Under the mixture model and the layout-smoothness condition,
\begin{equation}
    H_l^{(\mathrm{high})}(\wb)
    \leq
    \frac{\tau_l}{\lambda_{r,l}}H_l(\wb).
    \label{eq:main_tau_tail_summary}
\end{equation}
Therefore, $\tau_l<\lambda_{r,l}$ guarantees a nontrivial contraction,
with
$\kappa_l\geq1-\tau_l/\lambda_{r,l}>0$.
Under the stronger condition
$\tau_l<\lambda_{r,l}/2$, we further have
$\kappa_l>1/2$ and
$H_l^{(\mathrm{low})}>H_l^{(\mathrm{high})}$.
In this regime, the retained heterogeneity can be expressed as
\begin{equation}
    \delta_{\mathrm{high}}^2
    =
    \gamma\delta_{\mathrm{low}}^2
    +\delta_{\mathrm{res}}^2,
    \qquad
    \gamma
    :=
    \max_{l\in\mcal{S}_{\mathrm{filter}}}
    \frac{\tau_l}{\lambda_{r,l}-\tau_l}
    <1.
    \label{eq:main_strong_high_heterogeneity}
\end{equation}
}

\begin{corollary}[Uniform Partial Participation]
\label{cor:main_partial_participation}
Under the conditions of Theorem~\ref{thm:main}, {assume $K\geq2$, $p_k=1/K$, and $1\leq M\leq K$}. At every round, sample $M$ clients uniformly without replacement,
independently of the stochastic-gradient noise. The
server averages the $M$ returned models. Define
\begin{equation}
    q_M:={\frac{K-M}{M(K-1)}}.
\end{equation}
Then the bound in Eq.~\eqref{eq:main_convergence} remains valid after
replacing its last term by
\begin{equation}
    \frac{8}{(1-\rho)\sqrt{T}}
    \left(
        q_M\delta_{\mathrm{high}}^2
        +
        \frac{\sigma_{\mathrm{loc}}^2}{ME}
    \right).
    \label{eq:main_partial_participation_term}
\end{equation}
Thus unbiased partial participation preserves the
$\mathcal{O}(1/\sqrt{T})$ stationarity rate. When {$M=K$}, this result
reduces to Theorem~\ref{thm:main}.
\end{corollary}

{
Corollary~\ref{cor:main_partial_participation} is stated for equal client
weights and simple averaging. This setting is consistent with our controlled
Dirichlet-partitioned image benchmarks, where each client contains the same
number of training samples, i.e., $n_k=n=N/K$, so the sample-size-weighted
aggregation reduces to uniform averaging over the active clients.
}

The proof, together with an extension to unequal client weights under
unbiased importance-weighted aggregation, is provided in Section~II
of the supplementary material. {For an illustrative equal-client-weight setting with $K=100$
and participation ratio $M/K=0.1$}, we have $M=10$ and
$q_M=1/11$. Hence limited participation introduces an explicit
finite-population sampling term, but this term depends on the retained
heterogeneity $\delta_{\mathrm{high}}^2$, rather than the unfiltered
client disagreement.

\paragraph{Discussion.}
{
Theorem~\ref{thm:main} establishes the standard
$\mathcal{O}(1/\sqrt{T})$ stationarity rate for non-convex
optimization. The parameter $\rho$ quantifies the fraction of the
global descent signal removed by low-frequency suppression, whereas
$\delta_{\mathrm{high}}^2$ measures the client heterogeneity retained
by the actual update operator.
Equation~\eqref{eq:main_derived_high_heterogeneity} further relates
the retained heterogeneity to the spectral concentration of client
disagreement through the factor $1-\underline{\kappa}$.
The mixture/layout analysis provides an interpretable sufficient
condition for such a reduction, with
$\tau_l<\lambda_{r,l}/2$ corresponding to the stronger regime where
low-frequency heterogeneity constitutes a strict majority.
}
}

\section{Experiments}

    \begin{table*}[ht]
    \centering
    \caption{{Test accuracy (\%) on CIFAR-10, CIFAR-100, and Tiny-ImageNet under varying Non-IID settings using ResNet-20 with 100 clients, a client participation ratio of 0.1, and a fixed communication budget of 300 rounds. Results are reported as mean $\pm$ standard deviation over three runs with different random seeds.}}
    \label{tab1:main_result1}
    \resizebox{0.95\linewidth}{!}{
    \begin{tabular}{lcccccc}
    \toprule
    \multirow{2}{*}{Method} & \multicolumn{2}{c}{CIFAR-10} & \multicolumn{2}{c}{CIFAR-100} & \multicolumn{2}{c}{Tiny-ImageNet} \\
    \cmidrule(lr){2-3} \cmidrule(lr){4-5} \cmidrule(lr){6-7}
     & $\alpha=0.6$ & $\alpha=0.1$ & $\alpha=0.6$ & $\alpha=0.1$ & $\alpha=0.6$ & $\alpha=0.1$ \\
    \midrule
    FedAvg   & 77.02 $\pm$ 0.79 & 58.40 $\pm$ 1.30 & 35.37 $\pm$ 0.89 & 28.88 $\pm$ 0.42 & 31.22 $\pm$ 0.24 & 25.82 $\pm$ 0.74 \\
    FedProx  & 77.01 $\pm$ 0.36 & 58.29 $\pm$ 0.85 & 36.60 $\pm$ 1.81 & 28.29 $\pm$ 0.72 & 31.94 $\pm$ 0.74 & 25.47 $\pm$ 0.76 \\
    FedDyn   & 82.18 $\pm$ 0.22 & 69.34 $\pm$ 1.62 & 42.78 $\pm$ 1.47 & 34.54 $\pm$ 1.46 & 35.14 $\pm$ 0.12 & 30.32 $\pm$ 0.34 \\
    SCAFFOLD & 78.23 $\pm$ 0.66 & 56.00 $\pm$ 2.40 & 40.04 $\pm$ 0.67 & 28.83 $\pm$ 0.63 & 34.87 $\pm$ 0.62 & 27.18 $\pm$ 0.49 \\
    FedCM    & 84.03 $\pm$ 0.21 & 66.55 $\pm$ 1.73 & 38.25 $\pm$ 2.92 & 28.24 $\pm$ 0.92 & 22.59 $\pm$ 0.52 & 14.33 $\pm$ 1.44 \\
    FedSAM   & 75.88 $\pm$ 0.13 & 57.05 $\pm$ 0.51 & 35.59 $\pm$ 0.49 & 27.52 $\pm$ 0.39 & 29.56 $\pm$ 2.44 & 25.56 $\pm$ 0.11 \\
    FedDisco & 76.71 $\pm$ 0.66 & 58.40 $\pm$ 0.92 & 36.27 $\pm$ 1.25 & 27.87 $\pm$ 0.58 & 32.30 $\pm$ 1.24 & 25.76 $\pm$ 0.42 \\
    FedAWA   & 76.49 $\pm$ 0.55 & 58.11 $\pm$ 1.14 & 36.10 $\pm$ 1.05 & 28.12 $\pm$ 0.46 & 31.94 $\pm$ 0.91 & 25.65 $\pm$ 0.30 \\
    FedLWS   & 76.77 $\pm$ 0.80 & 58.29 $\pm$ 0.55 & 36.33 $\pm$ 1.28 & 27.84 $\pm$ 0.38 & 31.64 $\pm$ 0.06 & 25.73 $\pm$ 0.19 \\
    \textbf{Ours} 
             & \textbf{85.11 $\pm$ 0.11} 
             & \textbf{78.05 $\pm$ 1.88} 
             & \textbf{45.99 $\pm$ 1.12} 
             & \textbf{40.16 $\pm$ 0.64} 
             & \textbf{36.73 $\pm$ 0.38} 
             & \textbf{31.55 $\pm$ 0.19} \\
    \bottomrule
    \end{tabular}
    }
    \end{table*}

\subsection{Experimental Setups}

\textbf{Datasets.} We evaluate our methods on three widely used image classification benchmarks: CIFAR-10 \cite{cifar}, CIFAR-100 \cite{cifar}, and Tiny-ImageNet \cite{tinyimagenet}. These datasets vary in complexity, number of classes, and dataset size, providing a comprehensive evaluation of the robustness and generalization of federated learning algorithms. Specifically, CIFAR-10 contains 60,000 color images of size 32×32 pixels across 10 classes, with 50,000 training images and 10,000 test images. CIFAR-100 has the same total number of images but includes 100 classes, with 500 training images and 100 test images per class, making it a more challenging benchmark due to the increased number of classes and finer-grained categories. Tiny-ImageNet is a subset of the ImageNet dataset, consisting of 200 classes, each with 500 training images, 50 validation images, and 50 test images, with images resized to 64×64 pixels. Its larger number of classes and higher-resolution images present a significantly more challenging scenario for federated learning.

To simulate realistic data heterogeneity across clients, we adopt a Dirichlet distribution-based partitioning strategy. Specifically, for each dataset, we sample class distributions for each client from a Dirichlet distribution parameterized by $\alpha \in {0.1, 0.6}$. A smaller $\alpha$ value corresponds to more non-IID distributions, meaning that individual clients receive highly skewed subsets of classes, reflecting the imbalanced and biased data often observed in real-world federated scenarios. Conversely, a larger $\alpha$ value leads to more IID-like distributions, where each client’s local dataset more closely approximates the overall class distribution. To illustrate the severity of data heterogeneity under the highly non-IID setting, Figure 1 of the supplementary material visualizes the client-wise class distributions of CIFAR-10, CIFAR-100, and Tiny-ImageNet under Dirichlet partitions with $\alpha = 0.1$ and $\alpha = 0.6$. Each heatmap illustrates the proportion of samples per class for each client, where the horizontal axis corresponds to client indices and the vertical axis represents class IDs. As expected, the $\alpha = 0.1$ setting exhibits highly skewed and imbalanced data allocations: certain clients show dark, concentrated regions indicating dominance of only a few classes, while other classes may be nearly absent. This reflects a strongly non-IID scenario commonly observed in real-world FL applications. In contrast, when $\alpha = 0.6$, the heatmaps become noticeably smoother with reduced color variations, indicating that clients receive more balanced class distributions and the overall data heterogeneity is significantly alleviated. These two complementary visualizations allow us to systematically assess the behavior and robustness of different federated learning algorithms across varying levels of client data heterogeneity. Besides, we also use realistic FL dataset-FEMNIST \cite{fl_leaf}, medical imaging dataset-BloodMNIST \cite{medmnist} and NLP dataset-Yahoo! Answers as our benchmarks.

\textbf{Baselines.} We compare our approach with several representative federated learning algorithms, including: (1) \textbf{FedAvg}~\cite{fedavg}, the standard federated learning algorithm;
(2) \textbf{FedProx}~\cite{fedprox}, which introduces a proximal term to mitigate client drift under data heterogeneity;
(3) \textbf{FedDyn}~\cite{feddyn}, which employs dynamic regularization to correct local update bias;
(4) \textbf{SCAFFOLD}~\cite{scaffold}, which uses control variates to reduce gradient variance;
(5) \textbf{FedCM}~\cite{fedcm}, which enhances federated averaging by incorporating client-level momentum to correct local update bias and improve stability.
(6) \textbf{FedSAM}~\cite{fedsam1,fedsam2}, which integrates sharpness-aware minimization into federated optimization;
(7) \textbf{FedDISCO}~\cite{feddisco}, which optimizes federated aggregation by dynamically weighing local updates based on both dataset size and category distribution discrepancy;
(8) \textbf{FedAWA}~\cite{fedawa}, which optimizes aggregation weights based on client-level representations;
and (9) \textbf{FedLWS}~\cite{fedlws}, which applies a shrinkage operation on aggregation weights.

\textbf{Implementation Details.} We primarily adopt ResNet-20 \cite{resnet} as the backbone. Following \cite{fedsmoo, fedlesam}, all Batch Normalization (BN) layers are replaced with Group Normalization (GN) \cite{groupnorm} to enhance stability under data heterogeneity. We also evaluate ResNet-18, WideResNet-56\_2, DenseNet-121, and ViT as additional backbones. Unless otherwise specified, we use the FFT-based realization as the default implementation of SpecGradFilter.
For the federated learning process, we conduct 300 communication rounds with a client participation rate of 0.1. Each activated client performs 5 local epochs per round using an SGD optimizer. All methods share the same base protocol: 100 clients in total, active client ratio 0.1, batch size 50, and weight decay 0.001. {For the controlled Dirichlet-partitioned image benchmarks, we enforce an equal
number of training samples on every client, i.e., $n_k=n=N/K$, while using the
Dirichlet distribution only to control the client-wise class proportions.
Consequently, the sample-size-weighted aggregation in Algorithm~1 reduces to
uniform averaging over the $M$ active clients, and five local epochs with batch
size $B$ correspond to a fixed
$E=5\lceil n/B\rceil$
local stochastic-gradient steps per round.}

To ensure a fair and rigorous comparison, we carefully tune the learning rate and method-specific hyperparameters for every baseline via grid search. Specifically, we perform (at minimum) a 2D grid over the base learning rate
$
\mathrm{lr} \in \{0.01, 0.05, 0.1, 0.15\}
$
combined with 5 representative values of each method's specific parameter. For FedAvg and SCAFFOLD (which have no extra hyperparameters), we conduct an extensive 1D search over a wider learning rate range
$
\{0.01, 0.02, \dots, 0.2\}.
$
{If the best-performing configuration lies on the boundary of the
initial grid, we expand the search in the corresponding direction
until the performance decreases. Consequently, each reported
configuration is locally bracketed by lower-performing neighboring
settings within the expanded search grid.
}

The final hyperparameter configurations selected for each dataset are
reported in Table~III of the supplementary material. Each baseline is tuned
separately on each dataset using its corresponding search space. For a
given dataset, the selected configuration of the base optimizer is reused
unchanged when SpecGradFilter is added, so that the comparison isolates
the effect of spectral gradient filtering. Unless otherwise stated, the
same dataset-specific configuration is also used across the evaluated
heterogeneity levels and backbone architectures.

For SpecGradFilter, we use a fixed spectral filtering ratio of $r=0.05$
throughout all reported experiments. SpecGradFilter is applied to the
architecture-selected feature-extraction convolutional tensors of
ResNet-20, ResNet-18, DenseNet-121, and WideResNet; projection parameters,
normalization parameters, biases, and classifier heads are excluded.
For ViT, only the patch-embedding projection is filtered. Layer-selection
ablations are presented in Section~\ref{sec:layer-wise-ablation}.

\begin{table*}[t]
\centering
\caption{
Performance comparison of federated learning methods before and after integrating our method across different backbones on CIFAR-10 and CIFAR-100 with $\alpha=0.1$. {Results are reported as mean $\pm$ standard deviation over three runs with different random seeds.}
}
\label{tab2:main_result2}

\resizebox{0.8\textwidth}{!}{
\begin{tabular}{lcccccccccc}
\toprule

\multirow{2}{*}{Method}
& \multicolumn{5}{c}{CIFAR-10}
& \multicolumn{5}{c}{CIFAR-100} \\

\cmidrule(lr){2-6}
\cmidrule(lr){7-11}

& R20
& R18
& WRN
& Dense121
& ViT
& R20
& R18
& WRN
& Dense121
& ViT \\

\midrule
FedAvg
& {58.40 $\pm$ 1.30} & {71.15 $\pm$ 0.19} & {56.97 $\pm$ 0.28} & {65.94 $\pm$ 2.79} & {43.16 $\pm$ 0.16}
& {28.88 $\pm$ 0.42} & {38.15 $\pm$ 0.10} & {29.59 $\pm$ 1.34} & {34.69 $\pm$ 0.21} & {23.18 $\pm$ 0.10} \\

\textbf{+ Ours}
& \textbf{{78.05 $\pm$ 1.88}} & \textbf{{76.00 $\pm$ 0.18}} & \textbf{{76.54 $\pm$ 0.66}} & \textbf{{82.27 $\pm$ 3.72}} & \textbf{{45.78 $\pm$ 0.88}}
& \textbf{{40.16 $\pm$ 0.64}} & \textbf{{44.44 $\pm$ 0.06}} & \textbf{{50.45 $\pm$ 0.70}} & \textbf{{52.83 $\pm$ 1.10}} & \textbf{{24.52 $\pm$ 0.71}} \\

\midrule

FedProx
& {58.42 $\pm$ 1.44} & {71.50 $\pm$ 0.08} & {55.59 $\pm$ 1.86} & {66.09 $\pm$ 2.60} & {42.66 $\pm$ 0.07}
& {27.92 $\pm$ 0.87} & {37.53 $\pm$ 0.56} & {29.79 $\pm$ 0.53} & {34.99 $\pm$ 0.59} & {23.09 $\pm$ 0.32} \\

\textbf{+ Ours}
& \textbf{{78.56 $\pm$ 1.30}} & \textbf{{75.95 $\pm$ 0.06}} & \textbf{{76.78 $\pm$ 1.31}} & \textbf{{80.77 $\pm$ 2.59}} & \textbf{{45.75 $\pm$ 1.00}}
& \textbf{{40.50 $\pm$ 1.85}} & \textbf{{44.41 $\pm$ 0.36}} & \textbf{{50.97 $\pm$ 1.10}} & \textbf{{53.31 $\pm$ 0.40}} & \textbf{{24.55 $\pm$ 0.77}} \\

\midrule

SCAFFOLD
& {57.70 $\pm$ 0.18} & {73.43 $\pm$ 0.54} & {53.41 $\pm$ 1.44} & {66.43 $\pm$ 4.33} & {40.96 $\pm$ 0.45}
& {28.49 $\pm$ 0.71} & {43.43 $\pm$ 0.02} & {28.95 $\pm$ 0.61} & {39.84 $\pm$ 0.11} & {22.55 $\pm$ 0.49} \\

\textbf{+ Ours}
& \textbf{{79.30 $\pm$ 0.31}} & \textbf{{78.03 $\pm$ 0.20}} & \textbf{{78.17 $\pm$ 0.44}} & \textbf{{86.62 $\pm$ 0.91}} & \textbf{{43.20 $\pm$ 0.50}}
& \textbf{{46.81 $\pm$ 0.08}} & \textbf{{49.09 $\pm$ 0.33}} & \textbf{{57.52 $\pm$ 1.05}} & \textbf{{64.90 $\pm$ 1.22}} & \textbf{{24.23 $\pm$ 0.21}} \\

\midrule

FedAWA
& {58.69 $\pm$ 1.38} & {71.11 $\pm$ 0.33} & {55.00 $\pm$ 0.40} & {66.34 $\pm$ 4.54} & {42.66 $\pm$ 0.40}
& {28.14 $\pm$ 0.79} & {38.28 $\pm$ 0.05} & {29.46 $\pm$ 1.20} & {34.59 $\pm$ 0.54} & {23.09 $\pm$ 0.08} \\

\textbf{+ Ours}
& \textbf{{78.48 $\pm$ 0.73}} & \textbf{{75.94 $\pm$ 0.02}} & \textbf{{77.31 $\pm$ 0.63}} & \textbf{{82.27 $\pm$ 3.78}} & \textbf{{45.51 $\pm$ 0.33}}
& \textbf{{40.27 $\pm$ 1.65}} & \textbf{{44.69 $\pm$ 0.06}} & \textbf{{50.80 $\pm$ 0.22}} & \textbf{{52.44 $\pm$ 0.75}} & \textbf{{24.47 $\pm$ 0.64}} \\

\bottomrule
\end{tabular}
}

\vspace{-2mm}
\end{table*}

\subsection{Main Results}

\begin{figure*}[t]
    \centering
    \includegraphics[width=0.9\textwidth]{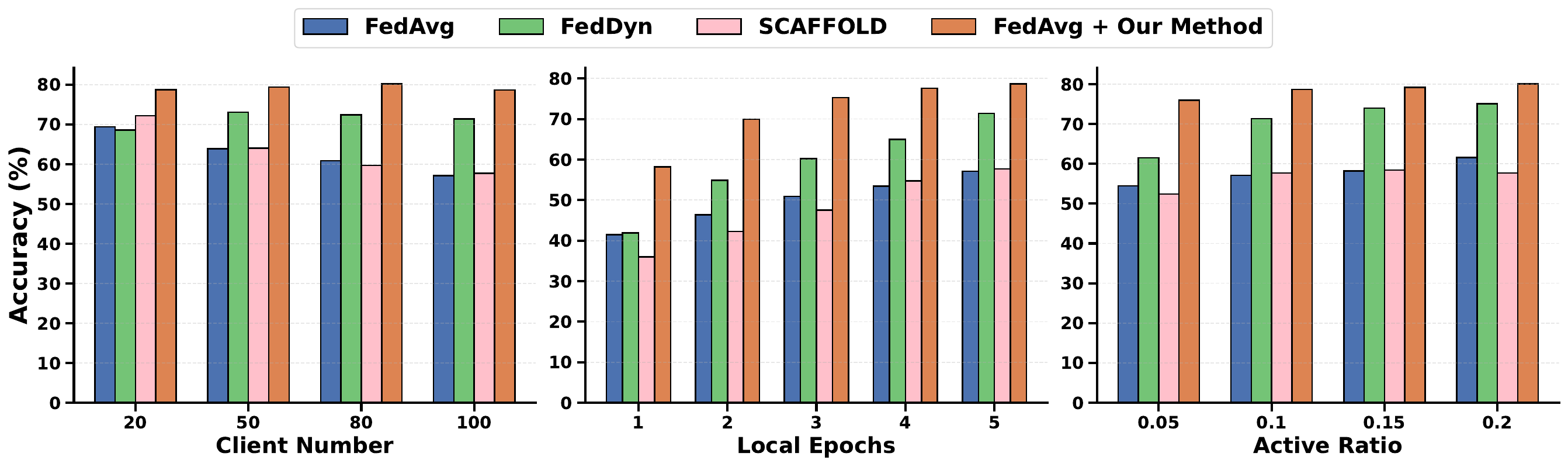}
    \caption{
    Comparison between our method and FedAvg on CIFAR-10 ($\alpha=0.1$)
    under different federated learning configurations. The default setting uses
    100 clients, 5 local epochs, and 0.1 participation rate, with each ablation
    varying only the corresponding hyperparameter.
    }
    \label{fig:fl_settings}

\end{figure*}

\begin{figure*}[t] 
    \centering
    \includegraphics[width=\linewidth]{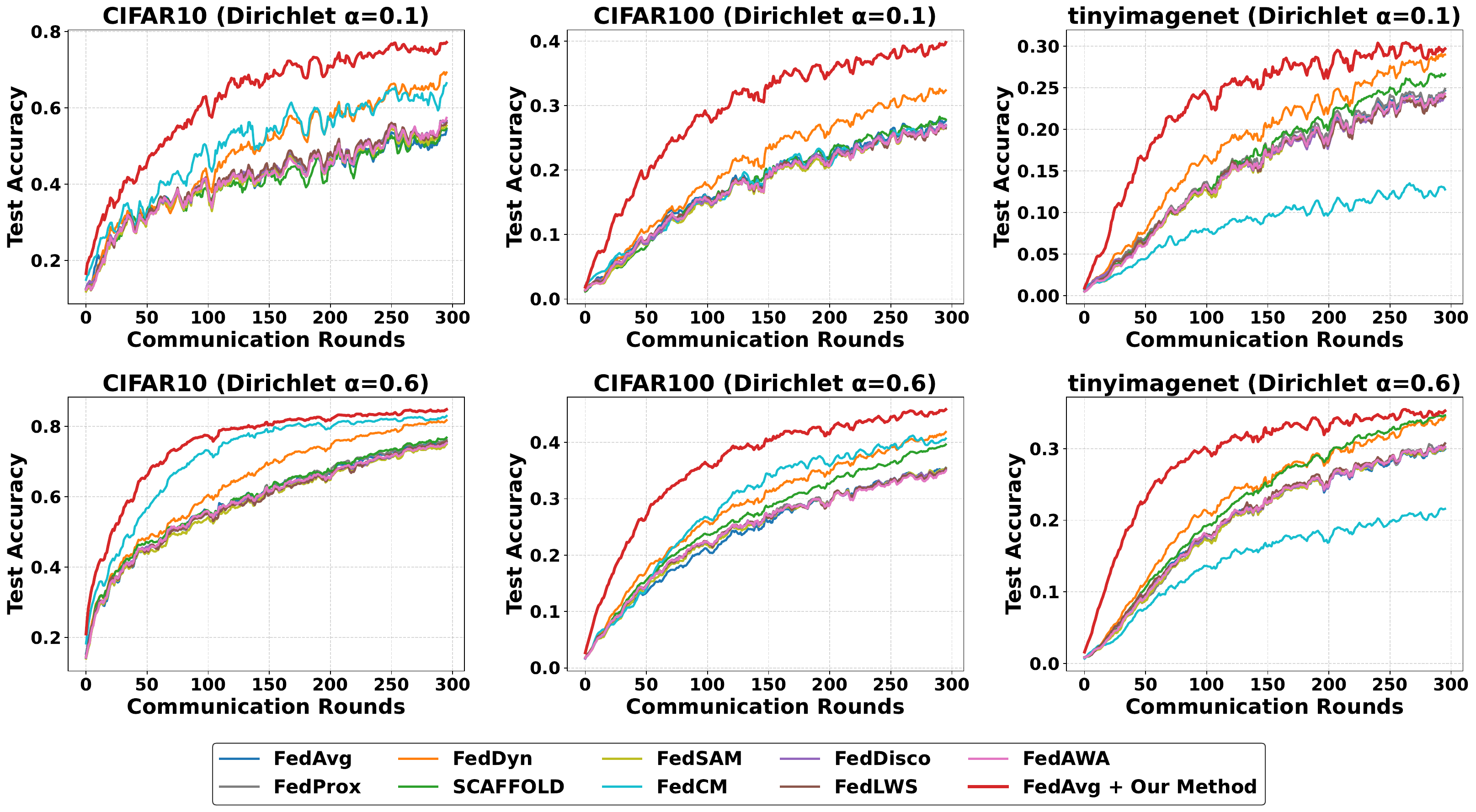}
    \caption{
    Test accuracy comparisons of different federated learning methods 
    across datasets and Dirichlet $\alpha$ settings on ResNet20. 
    }
    \label{fig:fl_all}
\end{figure*}

\textbf{Comparison with Baseline Methods.} Table~\ref{tab1:main_result1} reports the test accuracy under different degrees of data heterogeneity. Overall, our method consistently achieves the best performance across all datasets and Non-IID settings, demonstrating strong robustness to client drift. The performance gains are especially pronounced under severe heterogeneity ($\alpha = 0.1$) on CIFAR-10 and CIFAR-100, where our method significantly surpasses both FedAvg and strong drift-mitigation baselines. This trend aligns well with our empirical findings that client gradient divergence is dominated by low-frequency components under strong statistical heterogeneity, and that suppressing these components before local updates effectively improves update alignment across clients. Beyond final accuracy, our method also exhibits substantially faster convergence. As shown in  Figure~\ref{fig:fl_all}, it reaches high accuracy much earlier than competing methods, particularly on CIFAR-10 and CIFAR-100 with $\alpha = 0.1$, indicating that frequency-based gradient filtering not only stabilizes training but also accelerates global optimization within a fixed communication budget.

\begin{figure*}[t] 
    \centering
    \includegraphics[width=\linewidth]{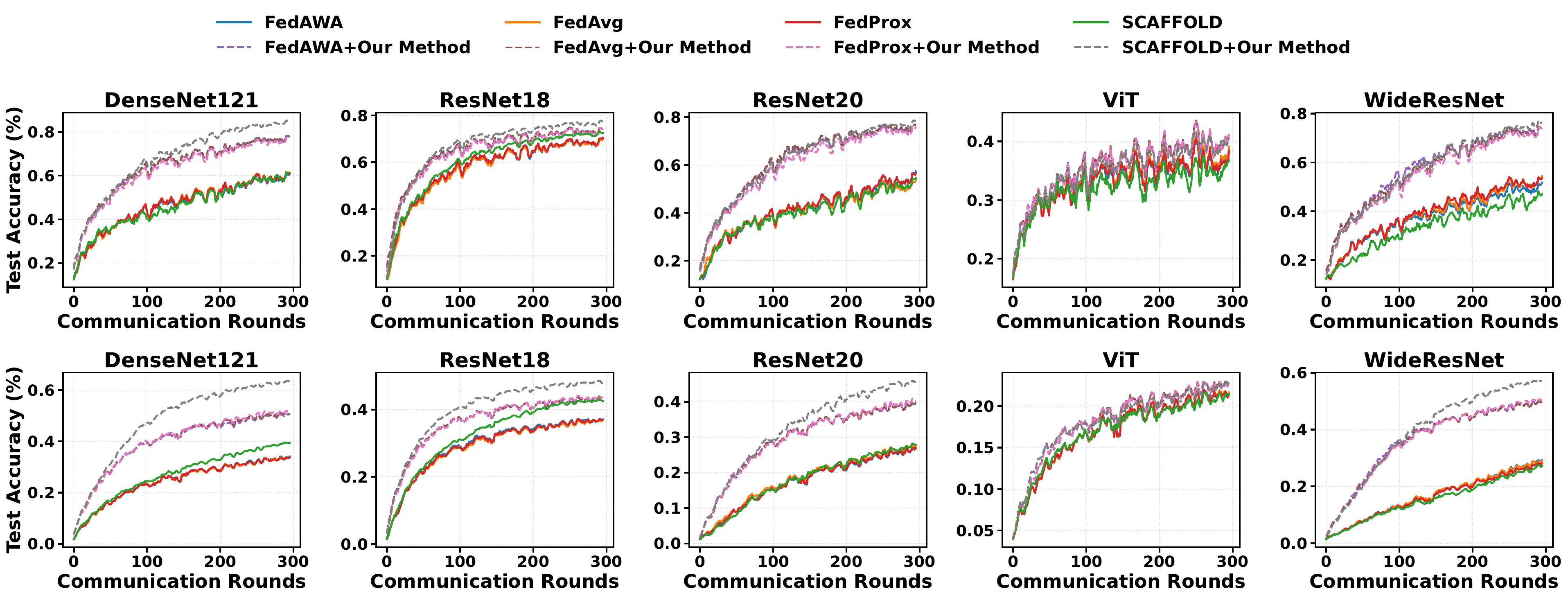}
    \caption{
    Test accuracy over communication rounds under Dirichlet ($\alpha = 0.1$). Solid lines denote the original federated learning methods, while dashed lines indicate the corresponding methods combined with our approach. The top row reports results on CIFAR-10, and the bottom row reports results on CIFAR-100. Each column corresponds to a different model architecture.
    }
    \label{fig:cross_model_learning_curve}
\end{figure*}

\textbf{Combination with FL Methods under Different Backbones.} Table~\ref{tab2:main_result2} reports test accuracy on CIFAR-10/100  under a highly heterogeneous setting ($\alpha=0.1$)  across different model backbones.  {For ViT, low-frequency suppression is applied only to the patch embedding gradients, as our layer-wise ablation study shows that filtering the patch embedding layer yields consistent improvements, whereas filtering other transformer layers provides limited or no performance gains.
} For FedProx and SCAFFOLD, the proposed filtering is applied only to the data-driven gradient components, while their method-specific correction terms (e.g., the proximal regularizer in FedProx and the control variate updates in SCAFFOLD) are left unchanged to preserve the original algorithmic structure. {The detailed integration procedures for different FL methods are provided in the supplementary material.} Ours consistently improves performance when integrated with various FL algorithms, across convolutional networks of different depths (ResNet20/18, WideResNet56\_2, DenseNet121) as well as transformer-based models (ViT). {These results demonstrate transfer across the tested architectures under architecture-specific layer selection. The gains on ViT are more moderate than most CNN gains and arise when filtering is restricted to the patch-embedding projection, consistent with the layer-wise ablation below.} On CIFAR-100, {where the classification task is more fine-grained}, the improvements are particularly significant for high-capacity models, supporting our motivation that suppressing client-specific low-frequency gradients leads to more stable and aligned updates. Furthermore, the learning curves in  (Figure \ref{fig:cross_model_learning_curve}) show that integrating ours consistently accelerates convergence, achieving substantially higher accuracy within the same communication rounds. {Collectively, these results show that SpecGradFilter can be integrated with the tested FL algorithms and backbones using the specified layer-selection and optimizer-integration rules.}

\subsection{Mechanism and Empirical Understanding}

\textbf{Ablation on Filtering Strategy.} To  validate our hypothesis that inter-client inconsistency is predominantly encoded in low-frequency gradient components, we conduct a comprehensive ablation study. We compare our proposed Low-Frequency Suppression strategies (including the FFT-based realization with a 5\% cutoff and the spatial variants LAP-D/GD) against counter-intuitive baselines: High-Frequency Suppression (removing the top 5\% frequencies), Random Suppression, and $\ell_2$-norm Rescaling (see Section I of the supplementary material for detailed settings). As shown in Table~\ref{tab:filtering_methods}, only strategies explicitly targeting low-frequency components yield substantial performance gains. Both LAP-D and GD significantly outperform FedAvg, confirming that removing smooth, client-specific gradient trends is critical for mitigating drift. In stark contrast, suppressing high-frequency components degrades performance, suggesting that high-frequency signals contain fine-grained discriminative features essential for generalization. Furthermore, the limited improvement of $\ell_2$-norm rescaling indicates that client drift is a structural spectral issue rather than merely a magnitude disparity. Notably, our FFT-based low-frequency suppression achieves the best overall accuracy on CIFAR-10 and remains highly competitive on CIFAR-100, empirically corroborating that low-frequency components are indeed the primary source of federated heterogeneity.

\begin{table}[t]
\centering
\small
\caption{Ablation of different filtering strategies on CIFAR-10 and CIFAR-100 under Dirichlet  ($\alpha$ = 0.1).}
\label{tab:filtering_methods}
\resizebox{0.8\linewidth}{!}{\begin{tabular}{lcc}
\toprule
\textbf{Filtering Strategy} & CIFAR10 & CIFAR100 \\
\midrule
None (FedAvg)              & 57.09 & 29.15\\
High-frequency suppression & 59.99 & 28.31 \\ 
Random suppression         & 58.89 & 28.74 \\
$\ell_2$-norm rescaling          & 59.33 & 27.04 \\
\midrule
Local Average Pooling Detrend  & 75.85 & 35.35 \\
Gaussian Detrend                   & \underline{77.06} & \textbf{41.53} \\
Low-frequency suppression  & \textbf{78.67} & \underline{40.87} \\
\bottomrule
\end{tabular}}
\end{table}

\textbf{{Layer-wise Filtering Ablation.}}
\label{sec:layer-wise-ablation}
{We investigate whether low-frequency filtering is equally effective across
different layer types. For ResNet-20, we separately filter the shallow
(\texttt{conv1} and \texttt{layer1}), middle (\texttt{layer2}), and deep
(\texttt{layer3}) convolutional stages, the classifier head, all
convolutional layers, and all convolutional layers together with the head. For ViT, we compare filtering the patch embedding, attention, MLP, classifier head, and all eligible weight layers. Normalization and bias
parameters are excluded, and all configurations use the same filtering ratio of $r=0.05$.
}

\begin{table}[t]
\centering
\caption{{
Layer-wise filtering ablation with ResNet-20 and ViT on CIFAR-10 and
CIFAR-100 under Dirichlet heterogeneity with $\alpha=0.1$.
Normalization and bias parameters are excluded.
}}
\label{tab:layerwise_filtering}
\setlength{\tabcolsep}{4.5pt}
\begin{tabular}{llcc}
\toprule
Backbone
& Filtered components
& CIFAR-10
& CIFAR-100 \\
\midrule
\multirow{7}{*}{ResNet-20}
& None (FedAvg)       & 57.09 & 29.15 \\
& Shallow Conv        & 60.46 & 30.28 \\
& Middle Conv         & 66.90 & 31.52 \\
& Deep Conv           & 73.89 & 36.81 \\
& Classifier Head     & 58.25 & 28.38 \\
& All Conv (Ours)     & \textbf{78.67} & \textbf{40.87} \\
& All Conv + Head     & 78.58 & 40.54 \\
\midrule
\multirow{6}{*}{ViT}
& None (FedAvg)          & 43.05 & 23.11 \\
& Patch Embedding (Ours) & \textbf{45.16} & \textbf{24.02} \\
& Attention              & 43.29 & 22.50 \\
& MLP                    & 42.63 & 22.92 \\
& Classifier Head        & 42.30 & 22.64 \\
& All Eligible Weights   & 44.77 & 23.26 \\
\bottomrule
\end{tabular}
\end{table}

{
As shown in Table~\ref{tab:layerwise_filtering}, the effectiveness of SpecGradFilter is strongly layer-dependent. For ResNet-20, the gain increases from shallow to deep convolutional stages, while filtering all
convolutional layers achieves the best performance. Filtering the classifier head alone provides no consistent benefit, and adding it to all convolutional layers does not further improve accuracy. For ViT, only the patch embedding yields consistent improvements, whereas filtering attention, MLP, or classifier parameters is ineffective.
These results indicate that SpecGradFilter is most effective on parameter tensors with convolution-like structural locality. {They also validate the layer selections used throughout our experiments: the architecture-selected feature-extraction convolutional tensors for CNNs and only the patch-embedding projection for ViT.}
}

\begin{figure}[t]
    \centering
    \includegraphics[width=\linewidth]{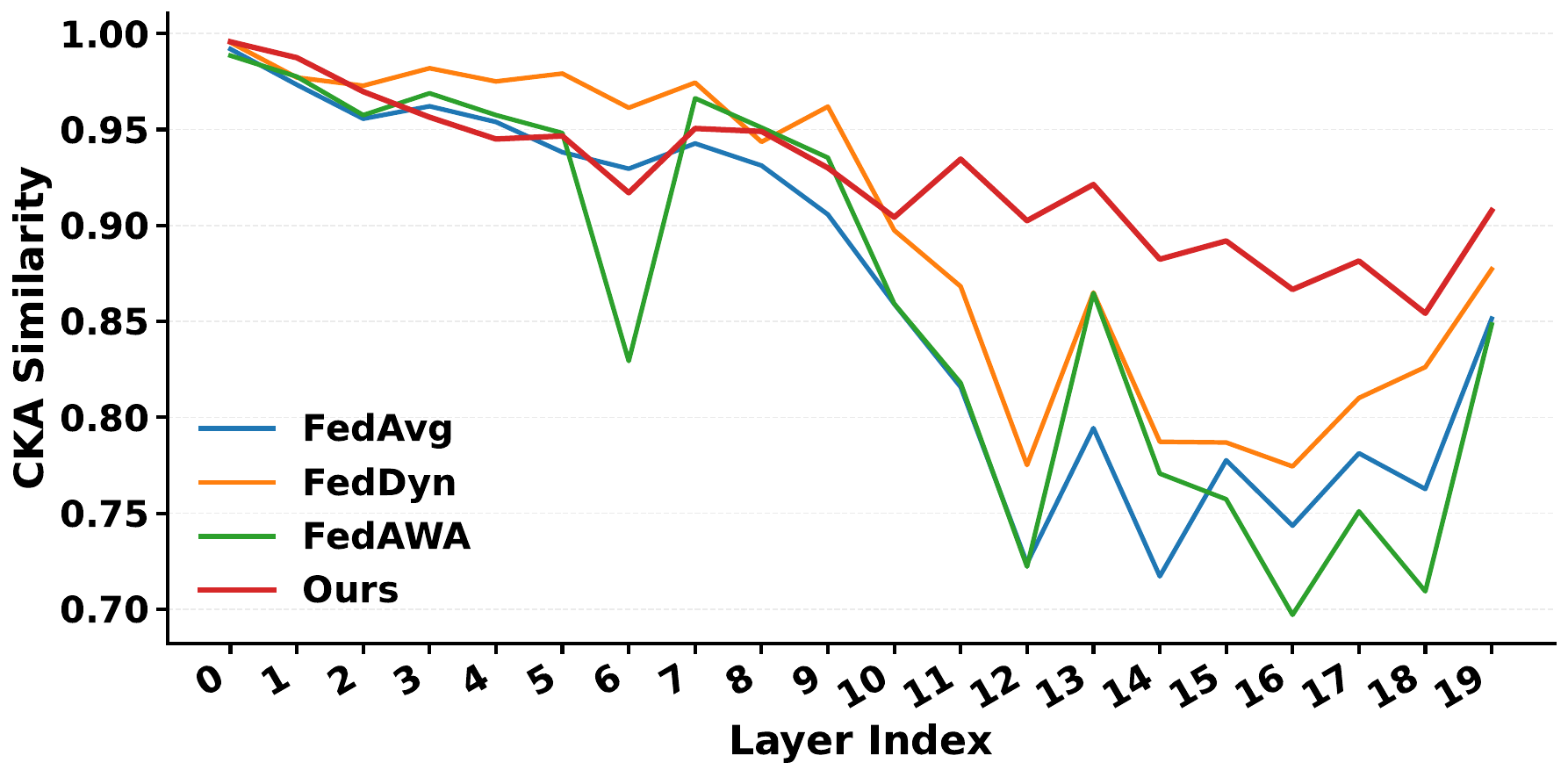}
    \caption{{
    Layer-wise CKA similarity among different methods on CIFAR-10 with
    $\alpha = 0.1$ using ResNet-20, showing improved representation alignment.
    }}
    \label{fig:cka}\vspace{-1em}
\end{figure}

\textbf{Representation Alignment via CKA.} To validate whether ours promotes structural consensus among clients, we visualize the Centered Kernel Alignment (CKA) similarity of the penultimate layer representations between pairwise local models (Figure~\ref{fig:cka}). While FedAvg exhibits low CKA scores (indicating that non-IID data drives clients to learn disjoint, distinct features), SpecGradFilter maintains consistently high CKA similarity across all layers. This suggests that by filtering out the low-frequency gradients which we identified as the carrier of client-specific bias, SpecGradFilter prevents the local models from diverging into incompatible feature spaces. Instead, it forces all clients to converge towards a shared, globally aligned representational structure, ensuring that the aggregated global model is constructed from semantically consistent local updates rather than conflicting ones.

\textbf{Why Does Flattening Order Matter?} Our default implementation performs a 1D FFT on flattened convolutional kernels following the natural contiguous tensor layout:
\[
[\text{Out}, \text{In}, \text{H}, \text{W}],
\]
which we denote as order \texttt{0123}. 
A natural question is whether the effectiveness of SpecGradFilter depends on this flattening strategy, or whether any arbitrary permutation would produce similar behavior.

To investigate this, we systematically evaluate multiple tensor flattening orders as well as a completely random permutation under the challenging Non-IID setting ($\alpha=0.1$). 
The results are summarized in Table~\ref{tab:flatten_order}.

\begin{table}[t]
\centering
\caption{Impact of flattening order on convolutional layers under severe heterogeneity ($\alpha=0.1$). Tensor dimensions follow the convention $[\text{Out}, \text{In}, \text{H}, \text{W}]$. {Results are reported as mean $\pm$ standard deviation over three runs with different random seeds.}}
\label{tab:flatten_order}
\resizebox{0.82\linewidth}{!}{
\begin{tabular}{lcc}
\toprule
Method & CIFAR-10 & CIFAR-100 \\
\midrule
FedAvg & {58.40 $\pm$ 1.30} & {28.88 $\pm$ 0.42} \\

0123 (ours) & \textbf{{78.05 $\pm$ 1.88}} & \textbf{{40.16 $\pm$ 0.64}} \\
0231 & {76.59 $\pm$ 0.28} & {38.83 $\pm$ 1.26} \\
0312 & {76.98 $\pm$ 1.75} & {38.09 $\pm$ 1.42} \\

3210 & {55.87 $\pm$ 0.44} & {26.16 $\pm$ 0.33} \\
1230 & {56.49 $\pm$ 0.96} & {25.85 $\pm$ 0.52} \\
2130 & {55.96 $\pm$ 1.03} & {25.87 $\pm$ 0.58} \\
2310 & {56.63 $\pm$ 0.49} & {25.57 $\pm$ 0.45} \\

2D FFT & {75.11 $\pm$ 0.25} & {31.30 $\pm$ 0.93} \\
Random after flatten & {57.06 $\pm$ 1.58} & {27.72 $\pm$ 0.89} \\
\bottomrule
\end{tabular}
}
\end{table}

\paragraph{Semantic Locality Matters}
Randomly permuting the flattened indices almost completely destroys the performance gain, reducing accuracy to a level comparable with or even worse than FedAvg. 
This observation suggests that SpecGradFilter does not function as a generic noise regularizer. 
Instead, its effectiveness critically depends on preserving meaningful structural locality in parameter space.

\paragraph{Why Do Natural Orders Work Better}
We hypothesize that client drift in federated learning primarily manifests as a macro-level cross-channel discrepancy rather than purely local spatial perturbations. 
Flattening orders such as \texttt{0123}, \texttt{0231}, and \texttt{0312} place the output-channel dimension near the outermost traversal order, causing inter-channel discrepancies to appear as slowly varying low-frequency patterns in the flattened spectrum.

Under this representation, client-specific drift becomes concentrated in low-frequency components and can therefore be effectively removed by high-pass spectral filtering, while fine-grained intra-kernel spatial structures remain preserved in higher frequencies.

In contrast, orders such as \texttt{2310} or \texttt{3210} place the output-channel dimension at the innermost traversal position. This rearrangement converts smooth cross-channel discrepancies into rapidly oscillating high-frequency patterns. As a result, the harmful drift is no longer captured in the low-frequency spectrum and therefore bypasses the spectral filter entirely, leading to performance even below FedAvg.

\paragraph{Comparison with 2D FFT}
We further implement a structure-preserving 2D FFT variant that independently applies spectral filtering to each spatial convolution kernel. Although the 2D FFT version still improves over FedAvg, it consistently underperforms our 1D SpecGradFilter, especially on CIFAR-100. This gap can be explained from two perspectives.

First, federated drift mainly occurs across channels and filters rather than within individual spatial kernels. 
The 2D FFT only analyzes isolated local $3\times3$ spatial structures and therefore cannot effectively capture inter-filter divergence. In contrast, our 1D flattening strategy concatenates filters into a unified spectrum, enabling direct modeling and suppression of global cross-channel drift.

Second, the 2D FFT implementation requires executing thousands of independent small FFT operations, introducing significant memory-access and looping overhead. 
Our 1D FFT instead operates on contiguous flattened arrays, resulting in substantially higher computational efficiency and simpler implementation.

Overall, these results demonstrate that the success of SpecGradFilter is tightly connected to the structural organization of neural parameters. 
Appropriate flattening orders preserve the semantic geometry of cross-channel optimization drift, enabling frequency-domain filtering to effectively separate transferable information from client-specific bias.

\textbf{When Does Low-Frequency Filtering Help or Hurt?} A natural question is whether suppressing low-frequency gradient components is always beneficial. 
To answer this, we investigate SpecGradFilter under different levels of decentralization and statistical heterogeneity, ranging from standard centralized training to highly heterogeneous federated learning. Our core hypothesis is that the role of low-frequency gradients fundamentally changes once optimization becomes decentralized.

\paragraph{Centralized Learning}
In standard centralized SGD, each mini-batch is sampled from the global data distribution, making the stochastic gradient an approximately unbiased estimator of the true optimization direction. 
Under this setting, low-frequency components largely encode globally consistent semantic structures and long-range optimization trends. 
Aggressively filtering these components therefore removes useful optimization signals and slows the learning of macro-level representations.

To verify this, we directly apply SpecGradFilter to centralized training. As shown in Table~\ref{tab:spectrum_heterogeneity}, low-frequency suppression slightly degrades performance compared with standard SGD (e.g., $-0.65\%$ on CIFAR-10 and $-1.57\%$ on CIFAR-100). 
This result confirms that when gradients are already globally consistent and unbiased, low-frequency filtering becomes unnecessary and may even be harmful.

\paragraph{Federated Learning}
The optimization dynamics fundamentally change in federated learning due to decentralized local updates. 
Under Non-IID data distributions, local gradients become heavily biased toward client-specific objectives. 
As a result, the low-frequency spectrum is no longer dominated by universal semantic information, but instead becomes contaminated by client-dependent drift and accumulated optimization bias.

In this regime, suppressing low-frequency components acts as a form of spectral drift correction, removing destructive global deviations while preserving transferable high-frequency micro-structures shared across clients. 
Consequently, the benefit of SpecGradFilter increases as heterogeneity becomes stronger. Table~\ref{tab:spectrum_heterogeneity} shows that the improvement consistently grows as the Dirichlet concentration parameter $\alpha$ decreases. 
For example, under severe heterogeneity ($\alpha=0.1$), our method improves FedAvg by $+21.58\%$ on CIFAR-10.

\paragraph{Why Does It Still Help Under Near-IID FL}
Interestingly, SpecGradFilter remains highly effective even under near-IID federated settings ($\alpha=100$), where global label distributions are nearly identical across clients. This phenomenon suggests that the effectiveness of spectral filtering is not solely caused by statistical heterogeneity. Even in near-IID FL, each client still performs multiple local SGD updates on different mini-batch sequences before synchronization. These local optimization trajectories accumulate stochastic variance and local overfitting tendencies, producing implicit low-frequency drift across clients.

Under this perspective, low-frequency filtering also behaves as a trajectory-level regularizer, suppressing unstable large-scale optimization directions and enforcing smoother inter-client consensus dynamics. As shown in Table~\ref{tab:main_result_single_col}, our method still substantially outperforms strong baselines such as FedDyn and SCAFFOLD under near-IID settings. Overall, these results reveal an important property of SpecGradFilter: its effectiveness is tightly coupled with decentralized optimization dynamics rather than merely data heterogeneity itself. When gradients are globally unbiased (centralized SGD), low-frequency suppression can remove useful semantic information. However, once optimization becomes decentralized, low-frequency components increasingly encode optimization drift and client-specific bias, making spectral filtering highly beneficial.

\begin{table}[h!]
    \centering
    \caption{Performance comparison across the spectrum of data decentralization and heterogeneity. In Centralized learning, the baseline is standard SGD; in Federated learning, the baseline is FedAvg. {Our spectral filtering slightly degrades performance in the tested centralized setting, whereas it improves the reported federated results, with larger gains at smaller $\alpha$.}}
    \label{tab:spectrum_heterogeneity}
    \resizebox{\linewidth}{!}{
    \begin{tabular}{l|ccc|ccc}
    \toprule
    \multirow{2}{*}{\textbf{Training Setting}} & \multicolumn{3}{c|}{\textbf{CIFAR-10 (\%)}} & \multicolumn{3}{c}{\textbf{CIFAR-100 (\%)}} \\
    \cmidrule{2-7}
    & Baseline & Ours & $\Delta$ & Baseline & Ours & $\Delta$ \\
    \midrule
    \multicolumn{7}{l}{\textit{Centralized Learning (Baseline: Standard SGD)}} \\
    Centralized (IID) & \textbf{92.87} & 92.22 & {-0.65} & \textbf{68.66} & 67.09 & {-1.57} \\
    \midrule
    \multicolumn{7}{l}{\textit{Federated Learning (Baseline: FedAvg)}} \\
    FL Near-IID ($\alpha = 100$) & 81.28 & \textbf{86.74} & \textcolor{increasegreen}{+5.46} & 40.90 & \textbf{49.36} & \textcolor{increasegreen}{+8.46} \\
    FL Non-IID ($\alpha = 5$)    & 79.58 & \textbf{86.77} & \textcolor{increasegreen}{+7.19} & 38.81 & \textbf{48.60} & \textcolor{increasegreen}{+9.79} \\
    FL Non-IID ($\alpha = 0.6$)  & 76.72 & \textbf{85.26} & \textcolor{increasegreen}{+8.54} & 36.09 & \textbf{46.66} & \textcolor{increasegreen}{+10.57} \\
    FL Non-IID ($\alpha = 0.1$)  & 57.09 & \textbf{78.67} & \textcolor{increasegreen}{+21.58}& 29.15 & \textbf{40.87} & \textcolor{increasegreen}{+11.72} \\
    \bottomrule
    \end{tabular}
    }
\end{table}

\begin{table}[ht]
        \centering
        \caption{{Test accuracy (\%) on CIFAR-10, CIFAR-100, and Tiny-ImageNet under near-IID ($\alpha=100$) setting using ResNet-20 with 100 clients, a client participation rate of 0.1, and a fixed communication budget of 300 rounds.}}
        \label{tab:main_result_single_col}
        \resizebox{0.89\linewidth}{!}{
        \begin{tabular}{lccc}
        \toprule
        Method & CIFAR-10 & CIFAR-100 & Tiny-ImageNet \\
        \midrule
        FedAvg        & 81.28 & 40.90 & 34.00 \\
        FedProx      & 81.62 & 40.56 & 33.73 \\
        FedDyn       & 85.42 & 46.45 & 37.69 \\
        SCAFFOLD   & 83.22 & 43.31 & 37.18 \\
        FedCM        & 86.36 & 43.86 & 23.56 \\
        FedSAM      & 81.42 & 39.50 & 34.23 \\
        FedDisco    & 81.30 & 40.00 & 33.25 \\
        FedAWA       & 81.51 & 40.32 & 33.93 \\
        FedLWS       & 81.52 & 39.73 & 34.75 \\
        Our Method  &  86.74 & 49.36 & 38.29 \\
        \bottomrule
        \end{tabular}
        }
\end{table}

\subsection{Robustness and Generalization}

\textbf{Robustness to FL Settings.}
We evaluate the robustness of our method under different federated learning hyperparameters, including the number of clients, local epochs, and client participation ratio, with FedAvg, SCAFFOLD, and FedDyn as baselines (Figure~\ref{fig:fl_settings}). 
As the number of clients increases from 20 to 100, FedAvg and SCAFFOLD exhibit noticeable performance degradation, with FedAvg dropping from 69.4\% to 57.1\%, indicating aggravated client drift under larger-scale heterogeneity. 
{In contrast, FedAvg combined with our method consistently maintains high accuracy across all displayed settings; FedDyn and SCAFFOLD are included as reference baselines in this figure.} 
{A similar advantage is observed when varying the number of local epochs or the client participation ratio, particularly under fewer local epochs and limited client participation.} 
Overall, these results demonstrate that frequency-based gradient filtering effectively stabilizes local optimization and improves robustness across different system configurations.

\begin{figure}[t]
    \centering
    \includegraphics[width=\linewidth]{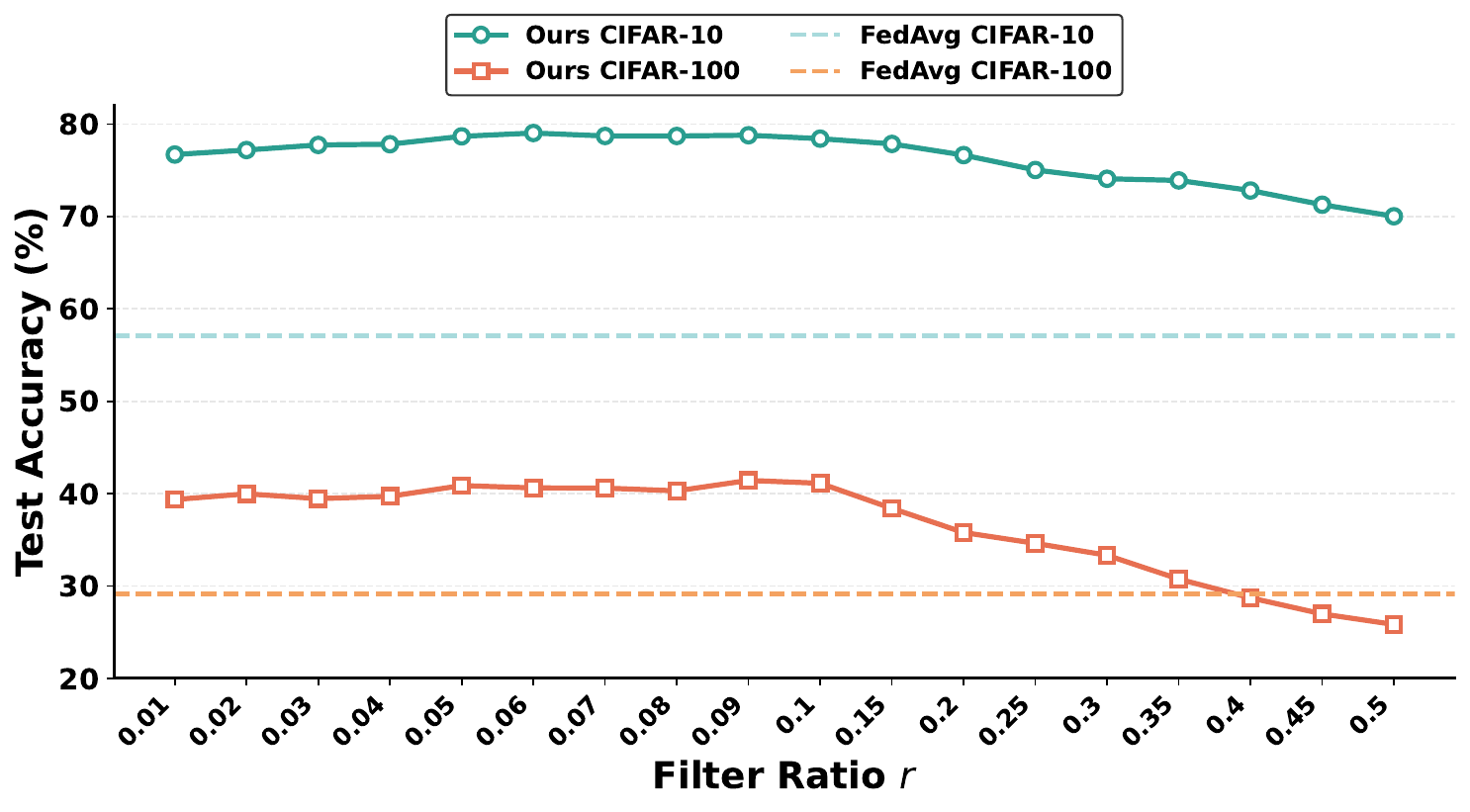}
    \caption{
    {
Test accuracy versus the filtering ratio on CIFAR-10 and CIFAR-100 under Dirichlet \(\alpha=0.1\) using ResNet-20. Dashed lines denote the corresponding FedAvg accuracies.
    }
    }
    \label{fig:different_ratios}
    \vspace{-2mm}
\end{figure}

{
\textbf{Impact of Filtering Ratio.}
We further investigate the sensitivity of SpecGradFilter to the filtering ratio on CIFAR-10 and CIFAR-100 under $\alpha=0.1$, extending the ratio from 0.01 to 0.50. As shown in Figure \ref{fig:different_ratios}, the performance remains relatively stable over moderate filtering ratios. On CIFAR-10, the accuracy stays within 76.70--79.03\% for ratios of 0.01--0.20, while CIFAR-100 maintains strong performance within 0.01--0.10, demonstrating that SpecGradFilter is robust to the filtering ratio within a reasonable range.

With larger ratios, the performance gradually decreases, particularly beyond 0.15 on CIFAR-100 and 0.20 on CIFAR-10. This is because excessive filtering may remove useful gradient information along with client-specific biased spectral components. Nevertheless, SpecGradFilter still clearly outperforms FedAvg on CIFAR-10 even at a ratio of 0.50. Overall, moderate filtering achieves a favorable balance between mitigating client-specific bias and preserving useful gradient information.
}

\begin{table*}[t]
\centering
\caption{
{Training efficiency and peak GPU memory comparison on an NVIDIA Tesla A40 with ResNet-20 and a local batch size of 50. We report the average time per communication round ($T_{\mathrm{rnd}}$), the number of communication rounds ($R$), the total training time
($T_{\mathrm{total}}$) required to reach the target accuracy achieved by FedAvg, and the peak GPU memory consumption ($M_{\mathrm{peak}}$).
The target accuracies are 57\% on CIFAR-10, 27\% on CIFAR-100, and 25\% on Tiny-ImageNet. Peak GPU memory is measured in MB using
\texttt{torch.cuda.max\_memory\_allocated()} during local training. ``$\dagger$'' indicates failure to reach the target accuracy.}
}
\label{appendix:time}

\definecolor{methodgray}{RGB}{245,247,250}
\definecolor{methodblue}{RGB}{230,242,255}

\setlength{\tabcolsep}{4.5pt}
\scriptsize

\resizebox{\textwidth}{!}{%
\begin{tabular}{
l
ccc>{}c
ccc>{}c
ccc>{}c
}
\toprule

& \multicolumn{12}{c}{\textbf{Dataset}} \\

\cmidrule(lr){2-13}

\textbf{Method}
& \multicolumn{4}{c}{\textbf{CIFAR-10}}
& \multicolumn{4}{c}{\textbf{CIFAR-100}}
& \multicolumn{4}{c}{\textbf{Tiny-ImageNet}} \\

\cmidrule(lr){2-5}
\cmidrule(lr){6-9}
\cmidrule(lr){10-13}

& $T_{\mathrm{rnd}}$
& $R$
& $T_{\mathrm{total}}$
& $M_{\mathrm{peak}}$

& $T_{\mathrm{rnd}}$
& $R$
& $T_{\mathrm{total}}$
& $M_{\mathrm{peak}}$

& $T_{\mathrm{rnd}}$
& $R$
& $T_{\mathrm{total}}$
& $M_{\mathrm{peak}}$ \\

\midrule

FedAvg
& \textbf{5.51} & 300 & 1653.00 & \textbf{100.22}
& \textbf{5.30} & 255 & 1351.50 & \textbf{100.30}
& \textbf{24.89} & 270 & 6720.30 & \textbf{255.21} \\

FedProx
& 5.86 & 255 & 1494.00 & 102.82
& 5.83 & 262 & 1527.46 & 102.93
& 26.46 & 255 & 6747.30 & 264.94 \\

FedDyn
& 6.10 & 167 & 1018.70 & 104.07
& 5.80 & 180 & 1044.00 & 104.18
& 26.05 & 194 & 5053.70 & 269.44 \\

SCAFFOLD
& 5.76 & 299 & 1722.20 & 100.32
& 5.75 & 262 & 1506.50 & 100.43
& 26.73 & 251 & 6709.23 & 258.77 \\

FedSAM
& 8.09 & 271 & 2192.40 & 101.27
& 8.56 & 284 & 2431.04 & 101.38
& 36.02 & 293 & 10553.86 & 259.56 \\

FedCM
& 6.04 & 128 & 773.10 & 104.28
& 5.73 & 284 & 1627.32 & 104.36
& $\dagger$ & $\dagger$ & $\dagger$ & 263.92 \\

FedDisco
& 5.54 & 254 & 1407.20 & \textbf{100.22}
& 5.79 & 270 & 1563.30 & \textbf{100.30}
& 25.98 & 299 & 7768.02 & \textbf{255.21} \\

FedAWA
& 5.52 & 255 & 1407.60 & 100.75
& 5.48 & 262 & 1435.76 & 100.91
& 25.34 & 270 & 6841.80 & 258.34 \\

FedLWS
& 5.62 & 255 & 1433.10 & 101.28
& 5.33 & 270 & 1439.10 & 101.38
& 24.86 & 270 & 6712.20 & 257.99 \\

\midrule

\rowcolor{methodgray}
FedAvg + LAP-D
& 5.97 & 97 & 579.10 & 101.27
& 5.86 & 128 & 750.08 & 101.35
& 25.34 & 108 & 2736.72 & 257.89 \\

\rowcolor{methodgray}
FedAvg + GD
& 6.66 & 79 & 526.10 & 101.27
& 6.42 & 93 & 597.06 & 101.35
& 25.98 & 98 & 2546.04 & 257.89 \\

\rowcolor{methodblue}
FedAvg + FFT (Ours)
& 6.38 & \textbf{75} & \textbf{478.50} & 101.27
& 6.25 & \textbf{86} & \textbf{537.50} & 101.35
& 25.52 & \textbf{93} & \textbf{2373.36} & 257.89 \\

\bottomrule
\end{tabular}%
}

\end{table*}

\textbf{Time Efficiency.} Although gradient filtering introduces a {measurable} per-round computational overhead, it is overshadowed by the substantial reduction in communication rounds required {to reach the specified target accuracies}. We evaluate the practical efficiency by measuring the wall-clock time required to reach the target accuracy. As listed in Table~\ref{appendix:time}, despite the increase in local computation, SpecGradFilter {reaches these targets in fewer rounds and less total wall-clock time in the tested setting}.
{Across all three datasets, the FFT-based realization and spatial variants substantially reduce the number of rounds and total training time required to reach the target accuracy. For the FFT realization, the total time-to-target is reduced by approximately $3.45\times$, $2.51\times$, and $2.83\times$ on CIFAR-10, CIFAR-100, and Tiny-ImageNet, respectively.
} This result highlights a critical trade-off: the minor cost of spectral filtering is a worthwhile investment for effectively slashing the communication bottleneck in heterogeneous FL environments. In addition to the measured wall-clock times, we provide a theoretical perspective on the computational cost of our method. For each layer, flattening the gradient and performing FFT followed by low-frequency suppression and inverse FFT requires $\mathcal{O}(d \log d)$ operations, where $d$ is the number of parameters in the layer. This overhead is much smaller than the standard backpropagation, which dominates the computational cost. We further propose two lightweight alternatives: Local Average Pooling Detrend (LAP-D) and Gaussian Detrend (GD), both of which operate in $\mathcal{O}(d)$ time while approximating the effect of FFT-based high-pass filtering, making them highly efficient for large models or resource-constrained clients.

\textbf{Robustness Under Feature Shift.} Previous experiments mainly focused on label-distribution heterogeneity induced by Non-IID partitioning. 
To further evaluate the generalization ability of SpecGradFilter, we investigate a more challenging feature-shift federated setting using CIFAR-10-C and CIFAR-100-C. Unlike standard label skew, feature shift introduces systematic corruption patterns into the input space, such as blur, noise, weather effects, and contrast variations. 
These corruptions create strong client-specific domain styles, substantially altering the global feature statistics observed by local models.

We conduct experiments on CIFAR-10-C and CIFAR-100-C with corruption severity level 3 using a ResNet-20 backbone. 
To simulate a highly heterogeneous feature-shift environment, we construct a federated system with 100 clients and a low client participation ratio of 0.1. 
Each client is assigned only 2--3 corruption types, resulting in severe domain inconsistency across local training distributions. 
All methods are trained for 300 communication rounds, and results are averaged over three random seeds. The results are summarized in Table~\ref{tab:feature_shift}.

\begin{table}[t]
    \centering
    \caption{Performance under feature-shift federated learning on CIFAR-10-C and CIFAR-100-C (severity level 3) using ResNet-20. We report mean test accuracy (\%) over three random seeds under a challenging setting with 100 clients and client participation ratio $R=0.1$.}
    \label{tab:feature_shift}
    \resizebox{0.9\linewidth}{!}{
    \begin{tabular}{lcc}
    \toprule
    Method & CIFAR-10-C & CIFAR-100-C \\
    \midrule
    FedAvg              & 62.11 $\pm$ 0.78 & 26.18 $\pm$ 0.53 \\
    FedDyn              & 68.74 $\pm$ 1.13 & 31.83 $\pm$ 0.41 \\
    FedCM               & 76.86 $\pm$ 0.34 & 23.45 $\pm$ 1.18 \\
    SCAFFOLD            & 61.25 $\pm$ 1.01 & 25.95 $\pm$ 0.63 \\
    FedAvg + Our Method & \textbf{77.68 $\pm$ 0.42} & \textbf{37.04 $\pm$ 0.28} \\
    \bottomrule
    \end{tabular}
    }
\end{table}

Feature-shift federated learning introduces a fundamentally different form of heterogeneity compared with label skew. 
Instead of altering label distributions, corrupted domains mainly modify global appearance statistics and low-level feature distributions. 
These systematic domain-specific perturbations generate strong low-frequency biases in local optimization trajectories, causing severe inconsistency across clients.

Under this perspective, the low-frequency spectrum increasingly captures corruption-dependent domain styles rather than universally shared semantic information. 
SpecGradFilter effectively acts as a spectral domain-detrending operator that suppresses these client-specific low-frequency biases while preserving consensus semantic structures contained in higher-frequency components.

As shown in Table~\ref{tab:feature_shift}, our method consistently achieves the best performance under severe feature shift. 
The improvement becomes particularly significant on CIFAR-100-C, where our method outperforms FedAvg by more than $10\%$.
Notably, the gain further enlarges under the challenging regime of low client participation and highly fragmented corruption distributions, demonstrating that SpecGradFilter remains effective even when heterogeneity originates from feature-level domain shifts rather than label imbalance. These results suggest that the benefit of spectral filtering is not limited to conventional Non-IID label skew, but extends more broadly to decentralized optimization scenarios where client-specific low-frequency biases dominate local updates. {Additional experiments on NLP and medical imaging tasks, further demonstrate the generalization and robustness of SpecGradFilter; detailed results are provided in the supplementary material Section VII.}

\textbf{Realistic Federated Setting.}
This section evaluates the proposed method under a realistic federated learning setting, where only a fraction of clients participate in each communication round.
Table~\ref{tab:femnist} reports the test accuracy on FEMNIST under different client active ratios.
The results show that our method consistently improves performance over all baselines across different participation rates.
Notably, even under low client participation, the proposed approach maintains stable and reliable performance gains, demonstrating its robustness in realistic federated scenarios with partial and dynamic client availability.

\begin{table}[t]
  \centering
  \caption{Test accuracy (\%) on FEMNIST under different client active ratios.}
  \label{tab:femnist}
  \setlength{\tabcolsep}{6pt}
  \small
  \begin{tabular}{lcc}
    \toprule
    Method
    & Active Ratio = 0.05
    & Active Ratio = 0.1 \\
    \midrule

    FedAvg
    & 78.95 & 78.02 \\
    \textbf{+ Our Method}
    & \textbf{81.58} & \textbf{82.56} \\
    \midrule

    SCAFFOLD
    & 75.16 & 77.48 \\
    \textbf{+ Our Method}
    & \textbf{77.57} & \textbf{82.34} \\
    \midrule

    FedAWA
    & 79.00 & 78.46 \\
    \textbf{+ Our Method}
    & \textbf{81.54} & \textbf{82.03} \\

    \bottomrule
  \end{tabular}
\end{table}

\vspace{-3mm}
\section{Conclusion}
We present {SpecGradFilter}, a communication-efficient method to mitigate client drift in federated learning via spectral analysis of model gradients. {By treating gradients as structured signals, we empirically identify the {Spectral Bias of Drift}: in the evaluated trajectories, lower-frequency bands of selected, structurally ordered parameter tensors carry a disproportionate fraction of inter-client disagreement. SpecGradFilter suppresses a small portion of these bands during local updates. For its FFT-based realization, the convergence guarantee is conditional on retaining sufficient global descent signal and bounded post-filtering heterogeneity.} Experiments on diverse datasets and models show consistent gains over drift-mitigation baselines, with faster convergence and higher accuracy. {Despite these advantages, the current implementation has two main limitations. First, it adopts a fixed filtering ratio across clients and training rounds, which may not fully adapt to layer-dependent, client-dependent, or time-varying spectral heterogeneity. Second, although SpecGradFilter does not increase the communication payload, it introduces additional local FFT/rFFT computation and temporary spectral-buffer overhead.} Future work includes {adaptive layer-, client-, and training-stage-aware spectral filtering} and extending this framework to large-scale models such as LLMs, where heterogeneity and optimization instability are more pronounced.

\section{Acknowledgment}

This work of Liyang Yuan and Dandan Guo was supported in part by the National Natural Science Foundation of China (NSFC) under Grant 62306125. 
Z. Lin was supported by the National Natural Science Foundation of China (NSFC) under Grant 62276004, the Beijing Natural Science Foundation under Grant L257007, and the Beijing Major Science and Technology Project under Grant Z251100008425006.

Generative AI tools (e.g., ChatGPT) were used to assist with language refinement. All scientific content, methodology, experimental results, and conclusions were developed and verified by the authors.

\bibliographystyle{IEEEtran}

\bibliography{ref}

\begin{IEEEbiography}[{\includegraphics[width=0.8in,clip,keepaspectratio]{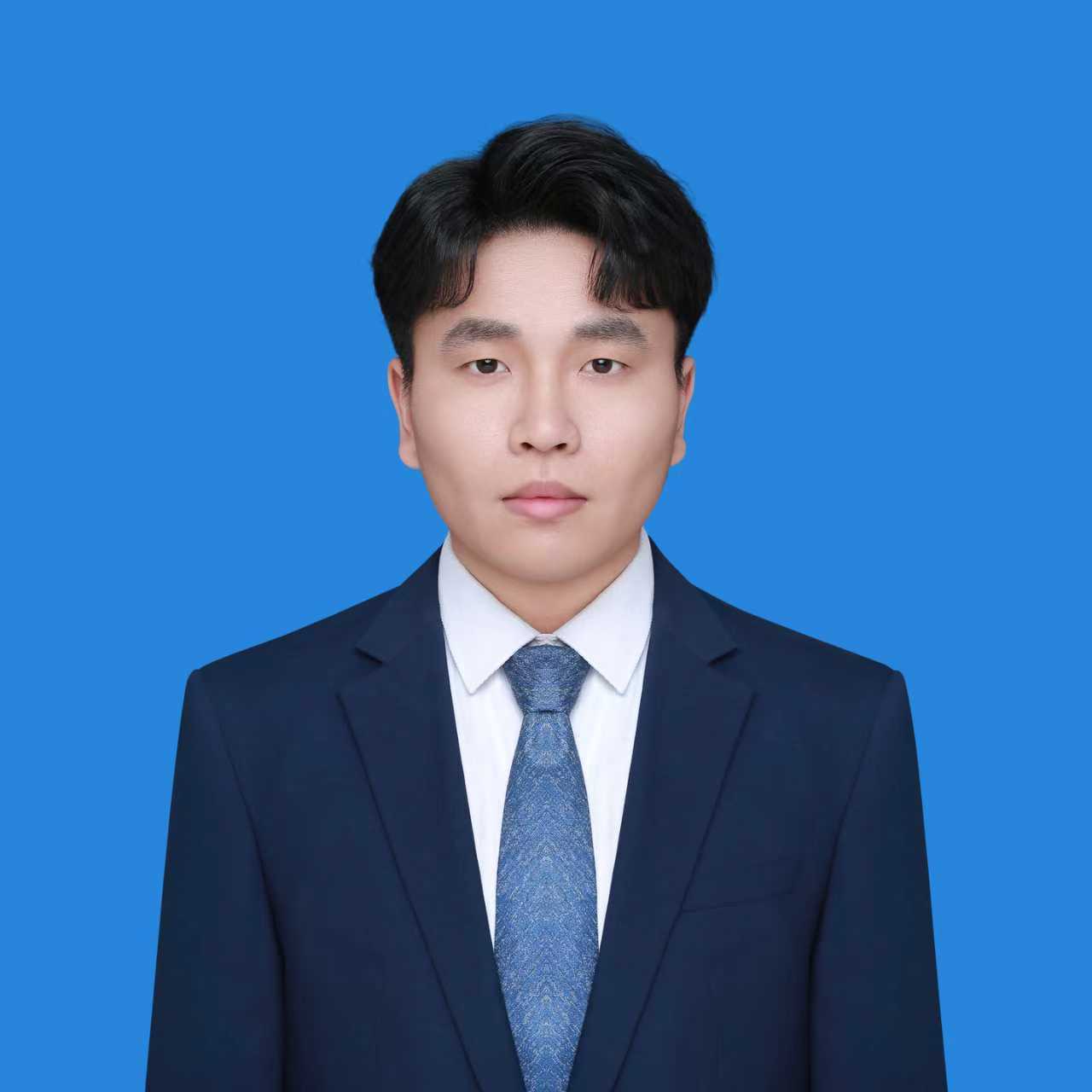}}]{Liyang Yuan} 
is currently working toward the M.S. degree with the School of Artificial Intelligence, Jilin University, Changchun, China. He received the B.S. degree in artificial intelligence from Ludong University, Yantai, China, in 2024. His current research interests include federated learning and client data heterogeneity.
\end{IEEEbiography}

\vspace{-3em}

\begin{IEEEbiography}[{\includegraphics[width=0.8in,clip,keepaspectratio]{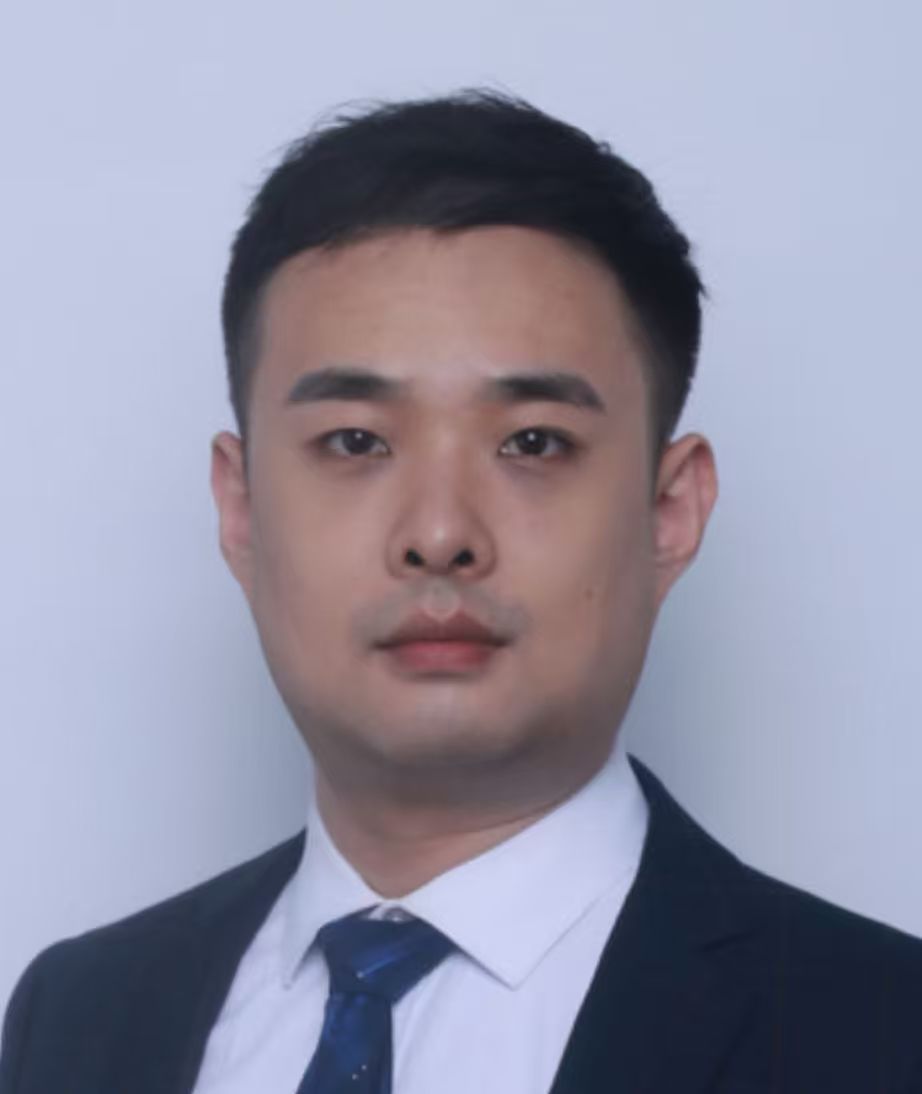}}]{Yibo Yang}
is a tenure-track associate professor with the School of Artificial Intelligence, Shanghai Jiao Tong University. He received his Ph.D. degree from Peking University and conducted research at KAUST and the University of Oxford. His research interests include foundation models and agents, lifelong learning, and trustworthy AI. His work has been published in leading venues including NeurIPS, ICML, ICLR, CVPR, IEEE TPAMI, and Science Advances. He has served as an Area Chair for NeurIPS, ICML, and ICLR, and as an Action Editor for TMLR.
\end{IEEEbiography}

\vspace{-3em}

\begin{IEEEbiography}
[{\includegraphics[width=0.8in,clip,keepaspectratio]{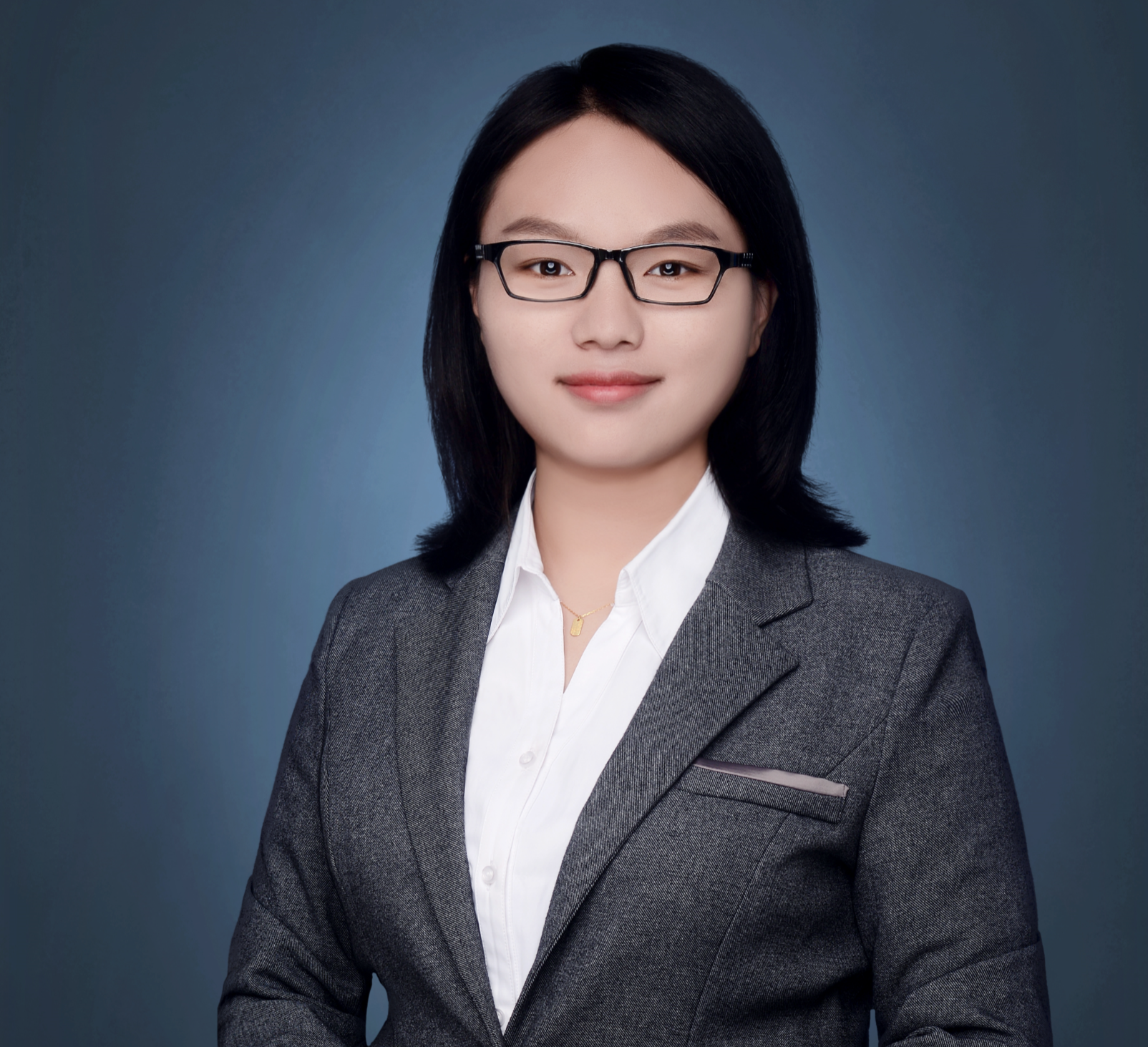}}]{Dandan Guo} is currently a Professor with the School of Artificial Intelligence, Jilin University, Changchun, China. Since 2025, she has also been a Visiting Faculty at King Abdullah University of Science and Technology (KAUST), Saudi Arabia. She received the Ph.D. degree in electronic engineering from Xidian University, Xi'an, China, in 2020. From 2020 to 2022, she was a Postdoctoral Researcher with The Chinese University of Hong Kong, Shenzhen, China. Her current research interests include advanced machine learning, federated learning, large language models (LLMs), and self-improving AI agents. 
\end{IEEEbiography}

\vspace{-3em}

\begin{IEEEbiography}
[{\includegraphics[width=0.8in,clip,keepaspectratio]{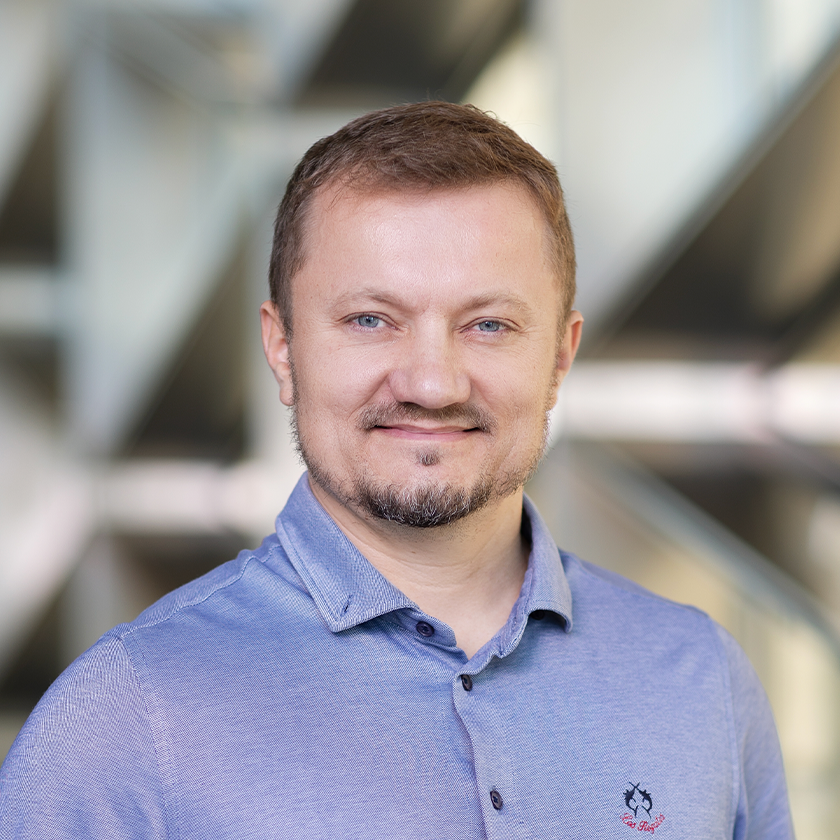}}]{Peter Richt\'arik} 
is a professor of Computer Science at KAUST, Saudi Arabia, where he leads the Optimization and Machine Learning Lab. Through his work on randomized and distributed optimization algorithms, he has contributed to the foundations of machine learning and optimization. He is one of the original developers of Federated Learning. Prof Richt\'arik's works attracted international awards, including the Charles Broyden Prize, SIAM SIGEST Best Paper Award, and a Distinguished Speaker Award at the 2019 International Conference on Continuous Optimization. He serves as an Area Chair for leading machine learning conferences, including NeurIPS, ICML and ICLR, and is an Action Editor of JMLR, and Associate Editor of Numerische Mathematik, and Optimization Methods and Software.
\end{IEEEbiography}

\vspace{-3em}

\begin{IEEEbiography}
[{\includegraphics[width=0.8in,clip,keepaspectratio]{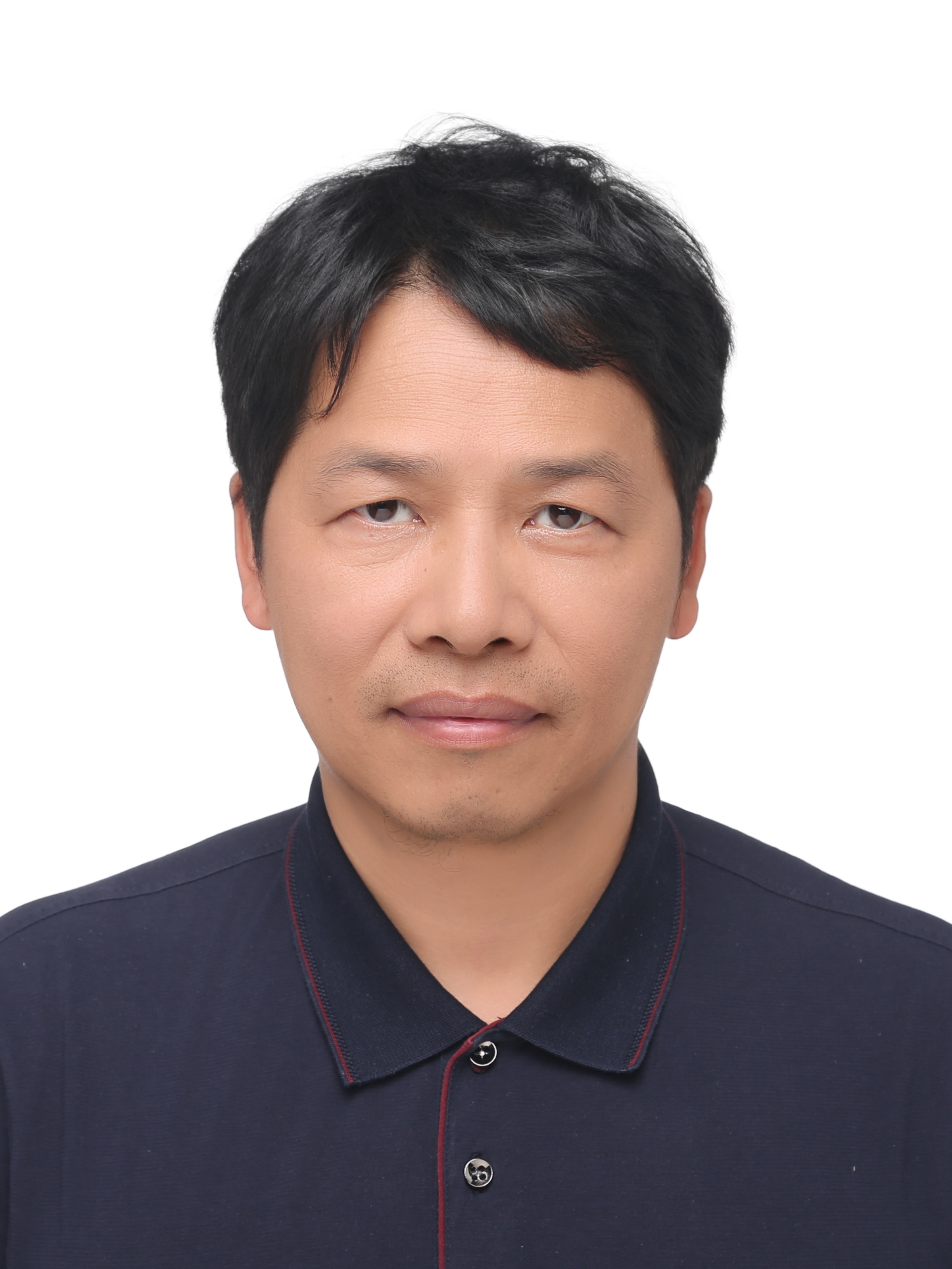}}]{Zhouchen Lin} (M’00–SM’08–F’18) is currently a Boya Special Professor with the State Key Laboratory of General Artificial Intelligence, School of Intelligence Science and Technology, Peking University. He received the Ph.D. degree in applied mathematics from Peking University in 2000. His research interests include machine learning and numerical optimization. He has published over 310 papers, collecting more than 45000 Google Scholar citations. He is a Fellow of the IAPR, the IEEE, the AAIA and the CSIG.
\end{IEEEbiography}

\end{document}